\documentclass[11pt]{article}

\usepackage[margin=1in, top=1in]{geometry}
\usepackage{fancyhdr}
\usepackage{titlesec}
\usepackage{titling}
\usepackage[most]{tcolorbox}
\tcbuselibrary{skins}
\usepackage{graphicx}
\usepackage{amsmath}

\usepackage{fontspec}
\usepackage{unicode-math}
\definecolor{qc_darkblue}{rgb}{0.008,0.063,0.247}
\definecolor{qc_blue}{rgb}{0.164,0.164,0.914}
\providecommand{\titlefont}{\sffamily\bfseries}

\titleformat{\section}{\sffamily\Large\bfseries\color{qc_darkblue}}{\thesection}{0.7em}{}
\titleformat{\subsection}{\sffamily\large\bfseries\color{qc_darkblue}}{\thesubsection}{0.6em}{}
\titleformat{\subsubsection}{\sffamily\normalsize\bfseries\color{qc_darkblue}}{\thesubsubsection}{0.5em}{}
\titlespacing*{\section}{0em}{1em}{.5em}
\titlespacing*{\subsection}{0em}{.7em}{.35em}

\newtcolorbox{titlebox}{
  enhanced, breakable, colback=white, boxrule=0pt, opacityback=0, opacityframe=0,
  width=\textwidth
}

\makeatletter
\newcommand{\contactinfo}[1]{\def\@contactinfo{#1}}
\makeatother
\contactinfo{}

\fancypagestyle{titlepage}{
  \fancyhf{}
  \fancyfoot[R]{\footnotesize\thepage}
  
}

\usepackage[numbers,sort&compress]{natbib}
\usepackage[colorlinks=true,linkcolor=qc_blue,citecolor=qc_blue,urlcolor=qc_blue,
            breaklinks=true,bookmarks=false]{hyperref}
\usepackage[capitalize]{cleveref}
\crefname{section}{Sec.}{Secs.}\Crefname{section}{Section}{Sections}
\crefname{table}{Tab.}{Tabs.}\Crefname{table}{Table}{Tables}
\crefname{figure}{Fig.}{Figs.}\Crefname{figure}{Figure}{Figures}
\crefname{equation}{Eq.}{Eqs.}\Crefname{equation}{Equation}{Equations}

\usepackage{booktabs,multirow,array}
\usepackage[font=small,labelfont={bf,sf},labelsep=period]{caption}
\usepackage[font=footnotesize,labelfont={bf,sf}]{subcaption}
\usepackage{enumitem}

\usepackage{wrapfig}
\usepackage{needspace}
\usepackage{pifont}  %
\usepackage{tikz}
\usetikzlibrary{arrows.meta,positioning,calc,fit,backgrounds,shadows.blur,%
            shadings,fadings,decorations.pathreplacing,decorations.markings,%
            shapes.geometric,shapes.misc,patterns}
\definecolor{tetherteal}{RGB}{18,120,120}
\definecolor{tethergray}{RGB}{90,90,90}
\definecolor{tetherink}{RGB}{28,36,44}
\definecolor{tetherbg}{RGB}{245,248,249}
\definecolor{tetheraccent}{RGB}{232,122,52}
\definecolor{tetherviolet}{RGB}{122,88,172}
\definecolor{tethersky}{RGB}{60,150,200}
\definecolor{tethershadow}{RGB}{198,204,210}
\definecolor{tethercard}{RGB}{255,255,255}
\definecolor{tetherregbg}{RGB}{252,237,234}
\definecolor{tetherregink}{RGB}{176,52,40}
\definecolor{f2peach}{RGB}{250,231,214}
\definecolor{f2blue}{RGB}{215,232,240}
\definecolor{f2pink}{RGB}{243,227,232}
\definecolor{f2lav}{RGB}{236,230,242}
\definecolor{f2cream}{RGB}{247,240,222}
\definecolor{f2pill}{RGB}{105,77,55}
\definecolor{f2teal}{RGB}{54,140,158}
\definecolor{f2maroon}{RGB}{158,59,82}
\definecolor{f2ink}{RGB}{60,54,48}
\tikzset{
  tcard/.style={rounded corners=6pt, draw=f2ink!25, line width=0.5pt, fill=white,
    inner sep=7pt,
    drop shadow={shadow xshift=0pt, shadow yshift=-1.1pt, shadow blur steps=9,
      shadow blur radius=2.4pt, opacity=0.20, fill=f2ink}},
  tframe/.style={rounded corners=4pt, draw=f2ink!18, line width=0.5pt, fill=white,
    inner sep=1.8pt,
    drop shadow={shadow xshift=0pt, shadow yshift=-1pt, shadow blur steps=9,
      shadow blur radius=2pt, opacity=0.20, fill=f2ink}},
  tstage/.style={rounded corners=6pt, draw=f2ink!25, line width=0.5pt, fill=white,
    inner xsep=9pt, inner ysep=7pt, text=f2ink, font=\small},
  tarrow/.style={-{Stealth[round]}, line width=2pt, draw=f2teal, line cap=round,
    line join=round, shorten >=1.5pt, shorten <=1.5pt},
  tloop/.style={-{Stealth[round]}, line width=1.6pt, draw=f2maroon,
    dash pattern=on 3pt off 2.4pt, line cap=round, line join=round,
    shorten >=1pt, shorten <=1pt},
  tbadge/.style={rounded corners=4pt, fill=f2pill, draw=none, text=white,
    font=\scriptsize\bfseries, inner xsep=6pt, inner ysep=2.5pt},
  thpill/.style={rounded corners=5pt, fill=f2pill, draw=none, text=white,
    font=\footnotesize\bfseries, inner xsep=9pt, inner ysep=3.5pt},
  tbadgered/.style={ tbadge, fill=f2maroon },
  tband/.style={rounded corners=9pt, fill=f2blue, draw=none, inner sep=11pt},
  tlabel/.style={ font=\footnotesize, text=f2ink },
}

\newcommand{\tether}{\textbf{Procedura}}

\newcommand{\runin}[1]{\vspace{2pt}\noindent\textbf{#1}\enspace}

\title{\includegraphics[height=4.2ex]{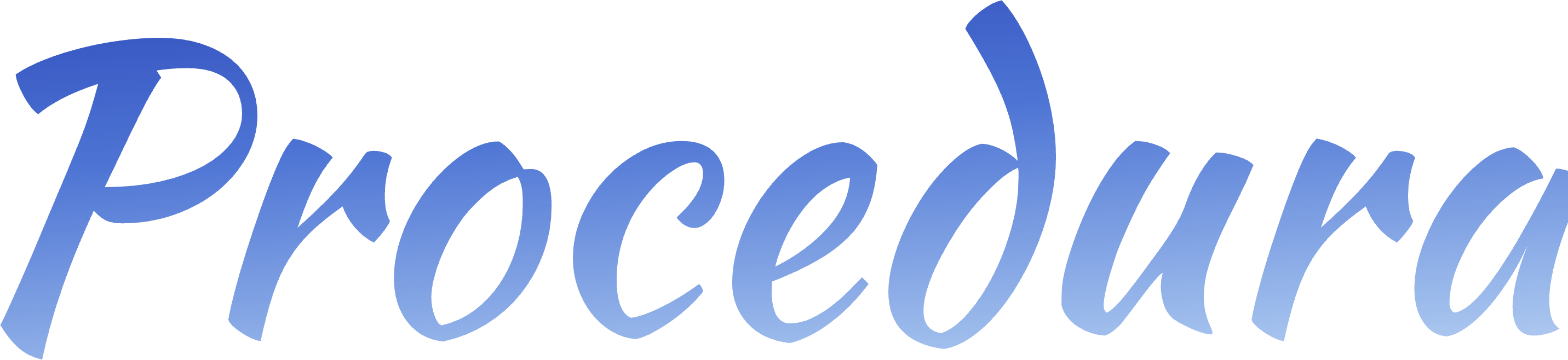}\\[0.15em]
Agentic 3D Modeling with Procedural Control}
\hypersetup{pdftitle={Procedura: Agentic 3D Modeling with Procedural Control}}
\author{Youtian Lin, Yikang Yang, Zhanpeng Hu, Mengqi Zhou, Feihu Zhang, Xun Cao, Jiaheng Liu, Yao Yao}

\begin{document}
\thispagestyle{titlepage}

\noindent\includegraphics[width=3.3cm]{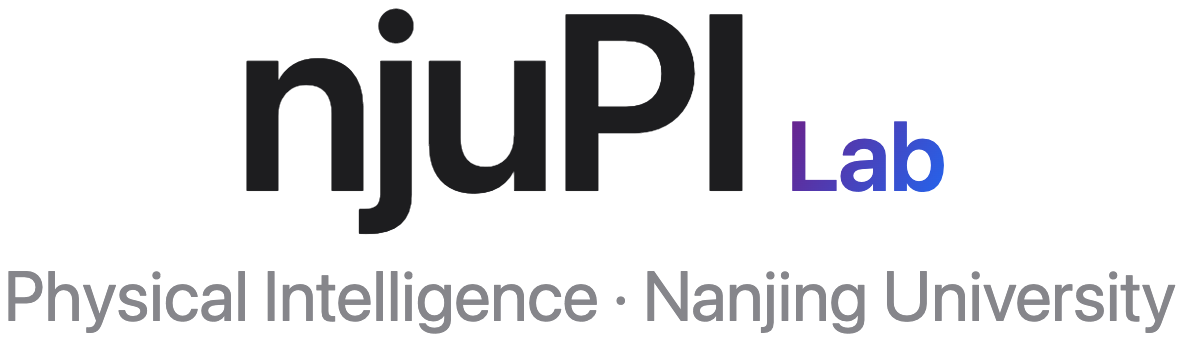}\par
\vspace{0.15em}

\begin{titlebox}
{\centering\titlefont\huge\bfseries\color{qc_darkblue}\thetitle\par}
\vspace{0.7em}
{\centering\sffamily\normalsize\color{qc_blue}%
  Youtian~Lin\textsuperscript{1}, Yikang~Yang\textsuperscript{1},
  Zhanpeng~Hu\textsuperscript{1}, Mengqi~Zhou\textsuperscript{1},
  Feihu~Zhang\textsuperscript{2}, Xun~Cao\textsuperscript{1}\\
  Jiaheng~Liu\textsuperscript{1,\dag},
  Yao~Yao\textsuperscript{1,\dag}\\[3pt]
  {\small\normalfont\itshape\color{qc_blue}\textsuperscript{1}Nanjing University
   \quad\textsuperscript{2}Envision \quad\textsuperscript{\dag}Corresponding authors\par}\par}
\vspace{0.45em}

\includegraphics[width=\linewidth]{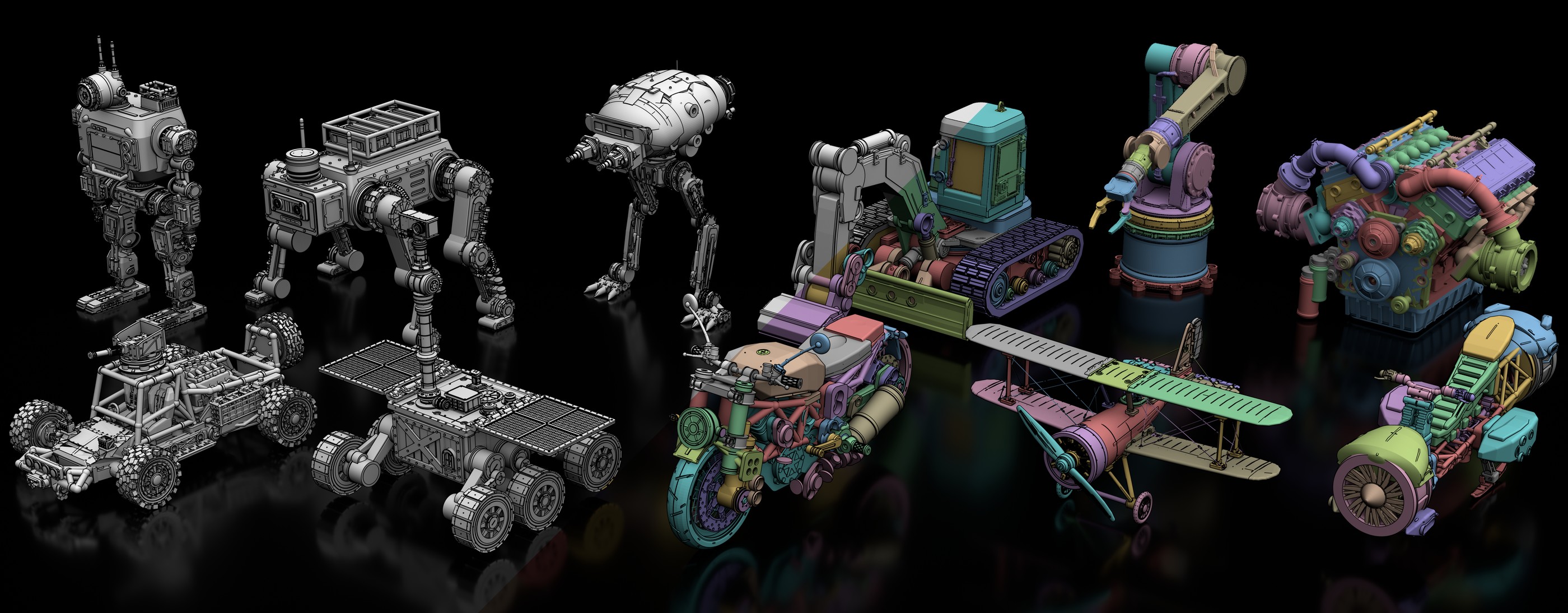}
\captionsetup{skip=4pt}
\captionof{figure}{\textbf{\tether{}} turns a text prompt into an editable procedural assembly, a parametric program whose named parts are joined by typed mates, written with a frozen LLM and no 3D training. Left: the compiled
geometry. Right: further objects coloured by their named modules, a part
decomposition that comes free with the representation.}
\label{fig:teaser}
\vspace{1.4em}

{Native 3D generators now recover impressive mesh geometry from a single image.
However, a dense mesh stays soft where a machined object should be sharp, it
carries no part decomposition, and it exposes no parameter a user could edit.
To address this, we explore the paradigm of 3D shape as code, leveraging and
scaling the coding ability of an LLM for 3D modeling. We introduce \tether{},
a novel 3D modeling agent framework that writes an object as a procedural
assembly, a parametric program whose named parts are joined by typed,
machine-checkable mates. From a text prompt, the agent plans the
object as an assembly graph and writes the program part by part, solving each
placement from the mated frames rather than guessing it, and admitting a part
only once compile, mate, and connectivity checks pass. A decoupled vision
critic then refines the assembly one diagnosed fix at a time. Moreover, the
same graph carries per-part materials and a simulator-validated articulation.
We evaluate on P3D-Bench under its assembly judge, and with the same judge on
MechBench-36, our hard-surface benchmark. On both, \tether{} outperforms
state-of-the-art native 3D generators and every prior 3D-code agent on judged
quality, produces the sharpest edges of any method we evaluate, and is the
only one whose output is an editable, part-structured program.
\par}
\end{titlebox}

\begin{figure}[t]
  \centering
  \includegraphics[width=\linewidth]{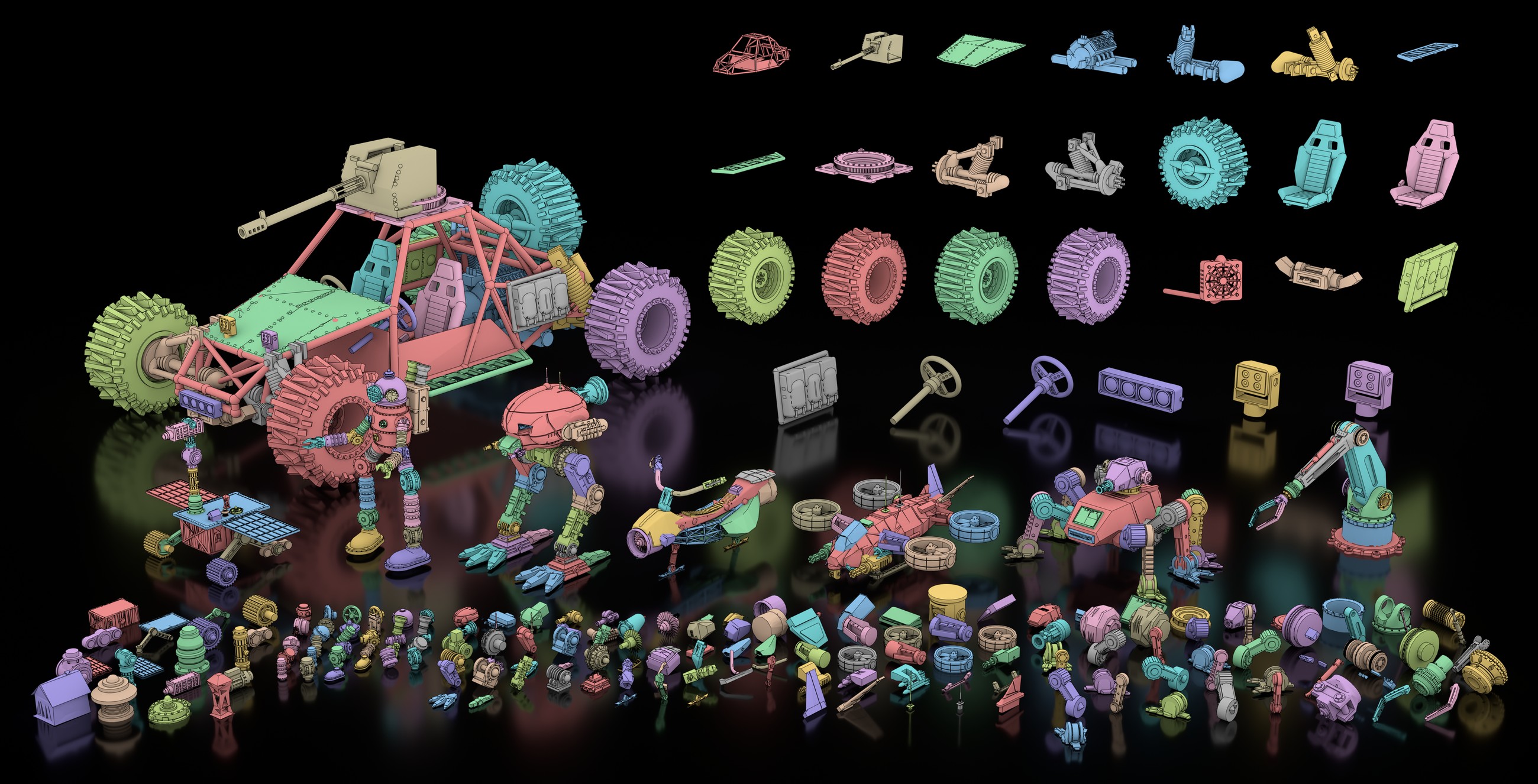}
  \caption{\textbf{A shape written as code is natively part-structured.} \tether{}
  builds each object as a parametric program with one named module per physical part,
  so the model decomposes into meaningful, editable segments at no extra cost, since
  there is no post-hoc segmentation step. The assault buggy (left) is shown with per-part
  colours and fanned out into its named parts (roll cage, turret, seats, wheels,
  drivetrain, panels), alongside other objects and a library of individual parts.
  A generated mesh is a single opaque surface. In contrast, every part here is
  individually addressable and re-parameterizable, and the same decomposition drives
  the per-part materials (\cref{sec:paint}) and articulation (\cref{sec:motion}).}
  \label{fig:parts}
\end{figure}

\section{Introduction}
\label{sec:intro}

\begin{figure}[t]
  \centering
  \includegraphics[width=\textwidth]{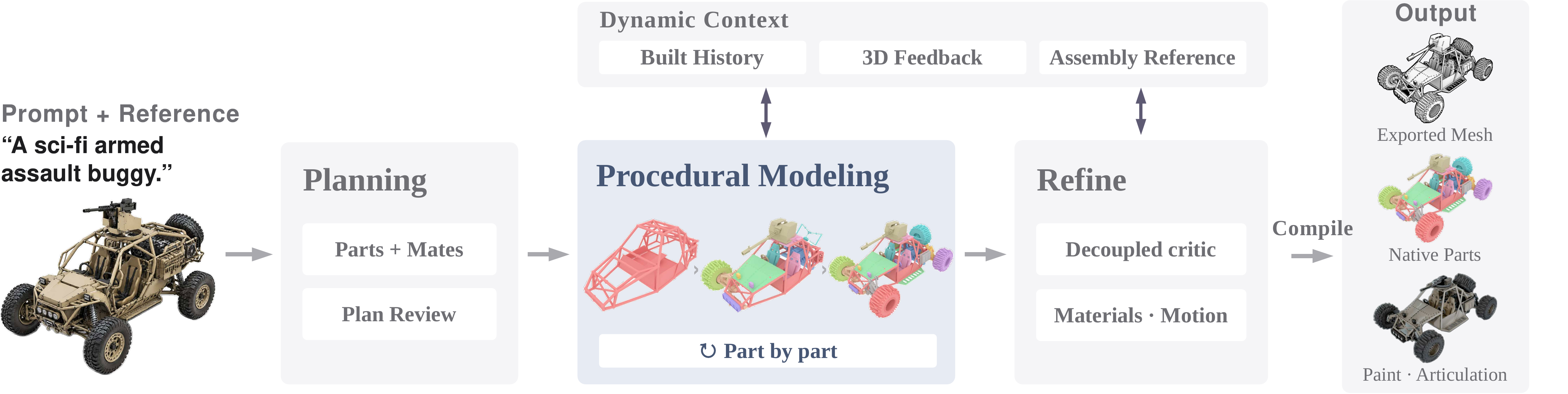}
  \caption{\textbf{Overview of \tether{}.} From a text prompt, \tether{} first
  synthesizes a reference view and then runs three stages. In Planning, it decomposes the object into an assembly graph of named parts joined by typed mates. In
  Procedural Modeling, the mate-driven build, it writes the 3D
  program part by part; each
  part is generated against its mates and a render of the partial build, its
  placement is solved from the mated frames, and a verification stack of compile, mate, and connectivity checks guards every commit. In Refine, a decoupled
  vision critic diagnoses the rendered model and gates one fix at a time. The
  same graph then completes the asset with per-part materials and an
  articulation validated in Isaac Sim.}
  \label{fig:pipeline}
\end{figure}

Machines, vehicles, and tools are the objects that fill games, simulators, and
factory floors, and generating them is a fundamental problem at the intersection
of computer vision and graphics. Native 3D generators have advanced rapidly and
now recover impressive mesh geometry from a single
image~\cite{wu2024direct3d,xiang2025trellis,hunyuan3d2025,jia2025ultrashape}.
However, neither of the two dominant generator families turns this fidelity
into a usable production asset. Native generators produce the shape as an SDF
or a voxel grid, whereas autoregressive generators write the mesh directly,
vertex by vertex and face by
face~\cite{nash2020polygen,siddiqui2024meshgpt,chen2024meshanything}. Both are
trained on large 3D datasets that are hard to collect, and both output raw
triangles with no structure attached. Neither family delivers geometry that is
both detailed and precise. Surfaces decoded from a grid stay soft and
scan-like where a machined object demands sharp edges and exact faces, and
meshes written token by token hold only as much detail as the token budget
allows. Nothing in the mesh says which triangles belong to which part. A part
decomposition must therefore come from a separate segmentation model applied
afterwards~\cite{yang2024sampart3d,liu2025partfield,yang2025holopart}, and
even then the cuts follow surface geometry rather than meaning, so we still
cannot reliably tell which piece is, say, the left finger of a robot hand.
Editing is also hard, since the mesh exposes no parameter to adjust; changing
the object means sculpting the surface directly, which shuts out the
non-expert user.
Taken together, prior methods fall short on high-precision geometry,
meaningful part decomposition, and editability.

Writing the shape as code offers another route, since a program is
parametric and part-structured by construction. State-of-the-art LLMs such as
GPT and Gemini~\cite{openai2023gpt4,gemini2023} now write high-quality code,
and this ability already reaches 3D, from Three.js scenes to Blender scripts~\cite{makatura2024scad,hu2024scenecraft}. Yet a 3D shape that an LLM writes as code
still comes out toy-like, extremely low-poly and without fine
detail~\cite{p3dbench}. The gap has a cause. A single pass must commit every
part, dimension, and placement at once, with no view of what its code
builds. Even recent agentic systems do not close this
gap~\cite{caddesigner2025,adamcad,zhou2026articraft}. Can we use the coding ability of an LLM to generate detailed, high-quality
3D shapes that meet the three requirements above?

To answer this question, we design \tether{} (\cref{fig:teaser}), an agentic
3D modeling framework that leverages the code generation ability of an LLM and
scales it to 3D shapes with high-grade detail and a native, meaningful part
decomposition. Specifically, \tether{} lets a frozen LLM write the shape in constructive
solid geometry (CSG), where every face is an exact plane or cylinder, so
edges come out sharp where a generated mesh stays soft. Our key idea is to
cast the code as a procedural assembly, a parametric program whose named
parts are joined by typed, machine-checkable mates such as a peg into a
socket or a hinge about an axis. This structure turns one large blind
generation into many small, checkable steps. Every part is a named, editable
module (\cref{fig:parts}) that the model writes in a single call, each
placement is solved from the mated frames rather than guessed, and compile,
mate, and connectivity checks admit every commit, so the build grows into a
detailed, many-part object without drifting~\cite{p3dbench}. Given a text
prompt, \tether{} synthesizes a reference view, plans the assembly graph,
builds the program part by part, and refines it with a decoupled vision
critic (\cref{fig:pipeline}).

We evaluate \tether{} on P3D-Bench under its assembly judge, and with the
same judge on MechBench-36, our hard-surface benchmark, against native
generators, prior 3D-code agents, and single-shot LLM baselines.
On both benchmarks, \tether{} outperforms all of them on judged quality.
Our contributions are as follows:

\begin{itemize}[leftmargin=1.2em,itemsep=2pt,topsep=3pt]
  \item \textbf{A 3D-shape-as-code modeling agent.} We explore the paradigm of
  3D shape as code and introduce \tether{}, an agentic framework that leverages
and scales the coding ability of a frozen LLM for 3D modeling, yielding
detailed, many-part 3D shapes from a text prompt with no 3D training.
  \item \textbf{Procedural assembly with typed mates.} We represent an
  object as a parametric CSG program whose named parts are joined by typed,
  machine-checkable mates. Placement is solved from the mated
frames and every commit is verified, so the build stays
connected and precise and every part remains editable by name. The same graph further carries per-part
  materials and a simulator-validated articulation.
  \item \textbf{Benchmark and state-of-the-art results.} On P3D-Bench and on MechBench-36, \tether{} outperforms state-of-the-art native 3D generators
  and all prior 3D-code agents, with the sharpest edges and the only editable,
  part-structured output.
\end{itemize}

\section{Related Work}
\label{sec:related}

\runin{Native 3D generation.}
A large body of work synthesizes a 3D shape directly as continuous geometry.
Optimization-based methods distill a 2D diffusion prior through score
distillation~\cite{poole2023dreamfusion,lin2023magic3d,wang2023prolificdreamer},
reaching high fidelity at the cost of a long per-object optimization.
Feed-forward methods map a prompt or a single image to a point cloud, an
implicit field, or a set of Gaussians in one
pass~\cite{nichol2022point,jun2023shape,zhao2023michelangelo,hong2024lrm,tang2024lgm},
and native 3D diffusion models now recover mesh geometry of striking fidelity
and broad
coverage~\cite{wu2024direct3d,wu2025direct3ds2,xiang2025trellis,hunyuan3d2025,jia2025ultrashape}.
In parallel, autoregressive models emit the mesh itself token by token~\cite{nash2020polygen,siddiqui2024meshgpt,chen2024meshanything}.
However, these models optimize fidelity to the input view alone. Trained on scanned and artist-made surfaces, they emit dense, soft, scan-like meshes with no named parts, no editable dimensions, and rounded edges where a machined object needs sharp ones. Part structure, when needed, comes from a post-hoc segmentation model~\cite{yang2024sampart3d,liu2025partfield,yang2025holopart}, which cuts the surface after the fact and attaches no dimensions or relations to the parts it names. In contrast, we
represent an object as the assembly program that constructs it, which makes
parts, dimensions, and relations explicit while requiring no 3D training.

\runin{Program-based 3D generation.}
Representing a shape as the program that builds it yields an inherently
editable, parametric model. Early work induces constructive solid geometry (CSG)
trees~\cite{sharma2018csgnet,kania2020ucsgnet} or structured assembly
programs~\cite{jones2020shapeassembly}, and CAD-native methods learn
sketch-extrude command sequences from large engineering
datasets~\cite{wu2021deepcad,willis2021fusion360,xu2022skexgen}. More recently, large language models write parametric 3D programs directly or translate text into construction-command sequences~\cite{makatura2024scad,khan2024text2cad}; semantic commenting then recovers part labels for an existing program~\cite{yuan2024cadtalk}. Production CAD treats assembly itself as a first-class object, where a
graph of typed mates between part-local connectors drives placement and
kinematics alike, and design-for-assembly practice prescribes how good parts
join: self-locating features, matched male--female pairs sharing one nominal
dimension, and explicit fits~\cite{boothroyd2010dfma}. Learned-assembly research
brings that structure to machine learning, predicting mates or joints for existing human-made parts~\cite{jones2021automate,willis2022joinable}
from large CAD datasets. However, no generative system authors this structure. The program-writing line remains single-shot, producing the code once with no visual feedback, so
many-part objects come out with wrong part counts and misplaced
parts~\cite{p3dbench}, and nothing in the emitted program states how two parts
relate. \tether{} generates the parts and their mate graph jointly,
building part by part against a render of what is already placed, so the structure that assembly research assumes as input becomes our output.

\runin{Agentic 3D-code generation.}
Closest to our setting, recent systems wrap a general LLM in a
draft-execute-revise loop that writes 3D code, runs the compiler, and revises from rendered or programmatic feedback~\cite{caddesigner2025,cadsmith2026,shui2026articad,zhou2026articraft,adamcad},
and domain-tuned coders fine-tune a model to emit CAD code~\cite{guan2025cadcoder,kolodiazhnyi2026cadrille}. Beyond static shape, agentic pipelines author articulated assets by
generating code for parts and their kinematics~\cite{zhou2026articraft,shui2026articad},
vision-language systems articulate an existing mesh by predicting its joints~\cite{le2024articulate}, and LLM scene systems arrange retrieved assets by predicted layout or
sample procedural
environments~\cite{feng2023layoutgpt,yang2024holodeck,hu2024scenecraft,deitke2022procthor,raistrick2023infinigen}. These systems
confirm that a coding agent can turn a prompt into an editable program, yet they share two limits. Parts are glued by raw transforms the model estimates from a picture, and kinematics is a stage that follows geometry, so a joint axis is
fitted to a finished surface rather than read off the structure that placed the
part. \tether{} answers both. Every relation
is a typed mate checked on the compiled mesh, so placement is
solved rather than guessed and connectivity is enforced during generation
instead of patched afterward~\cite{attene2013repair,huang2020manifoldplus};
kinematic mates live in the same graph, so a hinge is the very relation that
seated its two parts. Moreover, an independent vision critic, separate from the model that wrote the code, diagnoses every build and gates each edit. Self-refinement instead lets the same model critique its own output~\cite{madaan2023selfrefine,shinn2023reflexion,yang2023idea2img}, and such a reviewer often fails to find its own errors~\cite{huang2024selfcorrect}.

\section{Method}
\label{sec:method}

Given a text prompt, our goal is a 3D model that meets the three requirements
of \cref{sec:intro}: high-precision geometry, a meaningful part decomposition,
and editability. \tether{} therefore generates the program that builds the mesh rather than the mesh itself. We define this representation, a procedural
assembly whose parts are joined by typed mates (\cref{sec:rep}). The agent authors it in three stages (\cref{fig:pipeline}):
it plans the assembly graph (\cref{sec:plan}), builds the program one part at
a time under solved placement and deterministic verification
(\cref{sec:build}), and refines it through a decoupled vision critic
(\cref{sec:refine}). The same graph also completes the asset with per-part
materials (\cref{sec:paint}) and a simulator-validated articulation
(\cref{sec:motion}). One frozen LLM drives every reasoning step, so \tether{}
needs no 3D training; every other component in the pipeline is deterministic.

\subsection{Procedural Assembly Representation}
\label{sec:rep}

\runin{Formulation.}
\tether{} represents an object as a procedural assembly $A = (P, C)$, where
$P = \{p_1,\dots,p_n\}$ is an ordered set of parts, each a named parametric
CSG module, and $C$ is a set of mates, each a typed relation
between the mate frames of two parts. The shipped geometry is the mesh
$M = \textsc{compile}(A)$ that the compiler produces. This representation
answers the three requirements by construction. Geometry is precise since a program's faces are planes and its bores are cylinders, so its edges stay sharp at any resolution. The part decomposition is the module list itself, so it is exact and needs no separate segmentation model. Editing is a parameter change,
since every dimension is a named parameter bound to a named part. The representation also records relations. A mate states how two parts join, so placement can be solved and verified, and the same relation later drives articulation.

\begin{figure}[t]
  \centering
  \captionsetup{skip=3pt}
  \includegraphics[width=\textwidth]{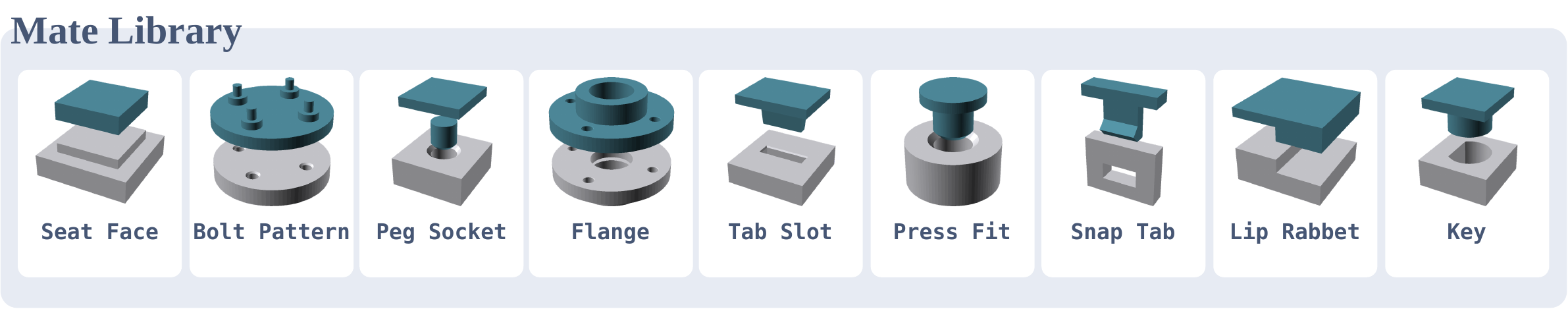}
  \caption{\textbf{The mate library.} The nine static mating
  features a part may declare, each rendered from the compiler macros the
  build emits. Teal is the male half added to the new part and grey the female
  half cut from its partner; both derive from one shared nominal dimension and
  a signed fit offset.}
  \label{fig:contracts}
  \vspace{-14pt}
\end{figure}

\runin{Typed mates.}
A mate $c = (i, j, \text{type}, F_i, F_j, d, \phi)$ joins a frame $F_j$
published by the new part $p_j$ to a frame $F_i$ published by an earlier part
$p_i$. Its type is drawn from a closed vocabulary of nine static mating
features (bolt-pattern, peg-socket, seat-face, flange, tab-slot, press-fit,
lip-rabbet, snap-tab, and key; \cref{fig:contracts}) and three kinematic ones
(revolute, prismatic, and spherical). The two halves share one nominal
dimension $d$, and a fit class $\phi \in \{$clearance, location, press,
snap$\}$ resolves to a signed per-side offset in millimetres. 
Two rules follow. First, both halves of a mate derive from the
same parameter $d$ plus a signed fit offset, never from two literals, so an
edit to one side can never orphan the other. Second, each type fixes a
degree-of-freedom signature, so a static mate fully locates its part while a kinematic mate leaves exactly the intended motion free.

\subsection{Assembly-Graph Planning}
\label{sec:plan}

\runin{Part plan with mates.}
Given the prompt, \tether{} first synthesizes a single reference view, which
grounds every later stage. A single vision call then decomposes the object
into an ordered list of named parts. Each part carries a \texttt{snake\_case}
module name, a coarse detail level (silhouette, major feature, or
sub-feature), and a short description that fixes its relative size, aspect,
and pose against its neighbours. Each part also carries its interface
declarations, the typed mates that bind it to parts placed earlier, each
naming the partner, the mate type, and the shared nominal. The list is ordered so that every part attaches to one placed earlier. It is therefore a topological order of the assembly graph, which lets the build seat each module on geometry that already exists. Mirror pairs, such as a left and
a right arm, are kept as separate parts, since a hand-assembled object is
rarely perfectly symmetric and later stages edit each side independently.

\runin{One-direction plan review.}
A second vision call reviews the plan against the reference, and we allow it to revise in only one direction. It may add a missing part or sharpen a description or mate, but it may
never merge, remove, rename, or reorder. The rule matters because each part
is exactly one generation call in the build, so merging two parts into one
removes detail the builder would otherwise produce. We enforce it in code
instead of trusting the prompt, so every original part survives with its name
and position, and only a genuinely new part is admitted, inserted after the
part it attaches to. The reviewed graph is the plan the build consumes.

\subsection{Mate-Driven Building}
\label{sec:build}

\begin{figure}[t]
  \centering
  \includegraphics[width=\textwidth]{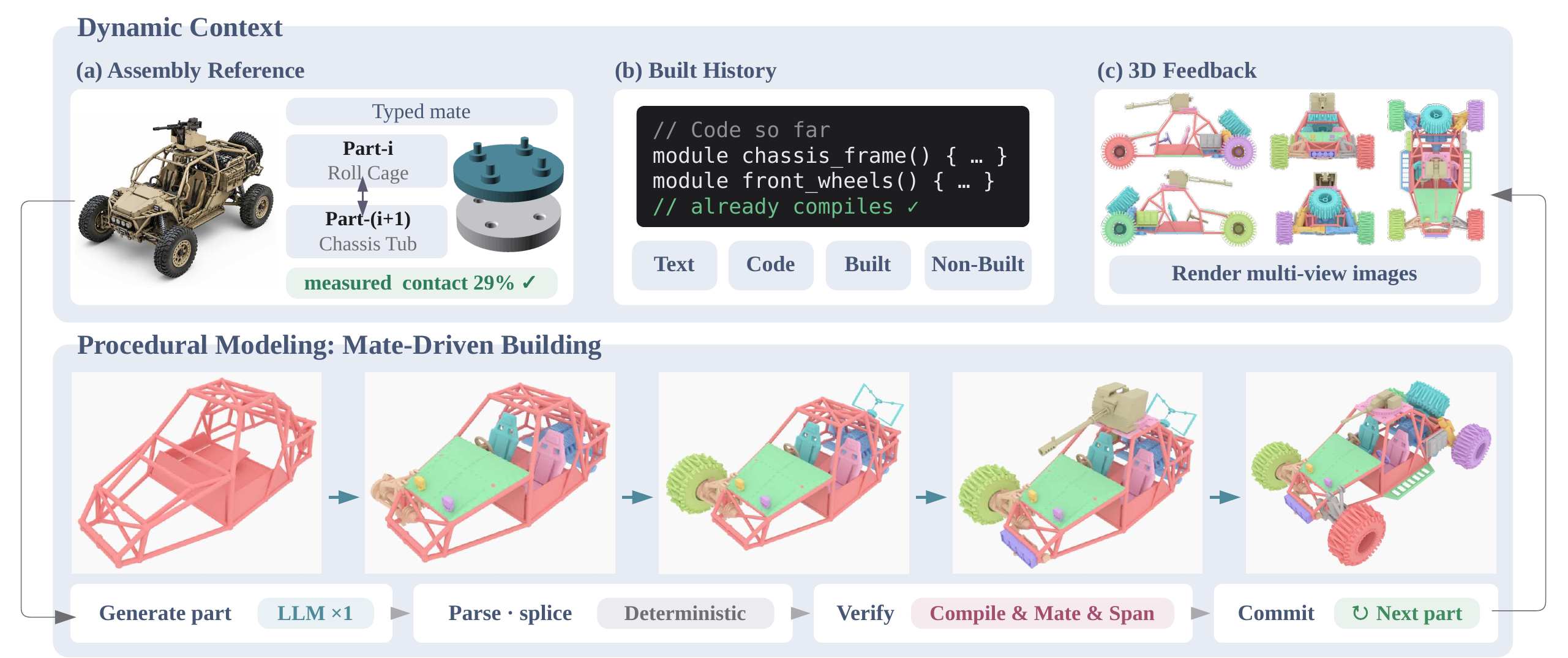}
  \caption{\textbf{Mate-driven building.} Top: for each planned part, the
  dynamic context packs three grounded sources into one generation
  message: (a) the assembly reference, namely the reference image, the prompt,
  and the part's mate, shown here beside the verdict its mesh check
  returned; (b) the built history, the full and already-compiling program; and
  (c) 3D feedback, parts-colour renders of the build so far.
  Bottom: the program
  grows one part per mate, shown as five snapshots of the assault buggy at
  $2$, $10$, $18$, $24$, and $28$ parts. Each part is generated in
  one LLM call and parsed into five blocks. The splice is deterministic and the
  placement is solved from the mated frames, and the compile, mate, and span
  gates must all pass before the part is committed. Every commit re-renders the
  3D feedback for the next part.}
  \label{fig:build}
\end{figure}

\runin{One part per call.}
A single LLM call can write the whole program for a simple shape. However, a
detailed, many-part object forces that one call to fix every part, count, and
placement at once, and the model loses track, leaving misplaced
sub-assemblies and floaters. To address this, the build grows the
program one module at a time, in plan order, so each call solves the small
problem of adding a single part to a model that already exists. The call
emits only the new part, in a strict five-block format: new parameters,
helper modules, the part body, its published interface frames, and its
placement intent. A simpler one-shot mode, which writes the entire program in
a single call, is the control our experiments compare against
(\cref{sec:exp}).

\runin{Dynamic context.}
Each generation call sees three grounded sources (\cref{fig:build}). The
assembly reference is the reference image, the text prompt, and the plan slice, namely the part to add, its mates, the parts not yet built, and a mandate
to match the reference's relative size, aspect, and pose. The built history
is the full accumulated program, passed as text; because it already compiles,
the model reads exact parameters and frames instead of guessing the current
scale. The 3D feedback is a parts-colour render of the build from up to
twelve viewpoints, six orthographic faces and six isometric corners. Each
committed part carries a distinct colour, so the model can see where its part
attaches. The first part, with nothing yet built, receives the reference and plan alone.

\runin{Solved placement.}
The model authors the part's geometry and its mate frames but not the assembly transform. For a static mate joining the new part's frame
$F_j$ to the placed partner frame $F_i$, the placement is the rigid transform
that aligns them:
\begin{equation}
  T_j \;=\; F_i \cdot \Delta(\phi) \cdot F_j^{-1},
  \label{eq:solve}
\end{equation}
where $\Delta(\phi)$ denotes the signed fit offset along the mate axis.
Solving $T_j$ in code removes the largest single source of drift in prior
3D-code systems, since the model never emits a raw transform it estimated
from a render. A part with several static mates is placed by the first
and checked against the rest. %

\runin{Deterministic splice.}
The five blocks are spliced into the accumulating program at fixed marker comments. Parameters are appended and de-duplicated, a
module whose name collides is renamed, and the solved placement is inserted
into the assembly. The LLM authors only the new part, so the program grows
without the generator ever re-emitting or corrupting what earlier calls
produced.

\runin{Verification stack.}
A spliced part must clear three deterministic checks before it joins the
build. First, a compile gate compiles the program; on failure the compiler's
error text is fed back and the part is regenerated. Second, a mate gate measures each of the part's static mates on the compiled mesh, testing
the registration area $a(c)$ of the mated surfaces and the penetration depth
$\delta(c)$ between the two bodies:
\begin{equation}
  a(c) \ge \tau_a \, d^2
  \quad\text{and}\quad
  \delta(c) \le \delta_{\max}(\phi) ,
  \label{eq:contract}
\end{equation}
where $d$ denotes the mate's nominal and $\delta_{\max}$ the tolerance
its fit class allows. A part that merely hovers near its partner, or gouges
into it, is therefore rejected, with the measured values returned in the
feedback. Third, a delta-aware connectivity gate reuses the span measure of
\cref{sec:refine} and rejects the part only when it increases the count of
visible floaters over the running baseline, so a part that legitimately
connects late is never blocked. Each part gets up to three quality attempts.
When it exhausts them it is committed with a warning rather than dropped,
since completion matters more than a defect the refine loop can still remove;
a part that never compiles is skipped. When the plan is exhausted, the
markers are stripped, yielding the draft assembly the refine loop takes up.

\subsection{Decoupled-Critic Refinement}
\label{sec:refine}

The draft assembly is complete but imperfect, so a refine loop polishes it.
Each cycle renders the current build, has a critic diagnose it against the
reference, and applies the top defect, followed by a verify compile. The defining
choice is to decouple the critic from the fixer, so the agent repairs what a
separate reviewer reports rather than what it judges of its own work.

\runin{Decoupled critic.}
The reviewer is a separate, single-turn vision call. It runs the same
underlying model with its own system prompt and context, never as a turn of
the agent that writes code. It sees the reference, the views the agent just
rendered, the parts-colour legend, and the current program, and returns a one-line
\texttt{SUMMARY} and a prioritized \texttt{ISSUES} list, each issue tagged
\texttt{[HIGH$|$MED$|$LOW]} and naming the responsible module with a
one-sentence fix direction. We decouple the two for three reasons. First, a model that judges
its own output carries the very assumptions that produced the error, so an
independent reviewer is more likely to catch it. Second, the diagnosis
becomes a persisted, first-class artifact, whereas in a freeform loop the
critique lives only in hidden reasoning. Third, the critic is history-aware, receiving a trail of what
prior cycles flagged, so it verifies that earlier fixes landed and flags
regressions
(\cref{fig:refineabl}).

\runin{Diagnosis-gated edits.}
The ordering is a hard invariant, enforced in code. A successful
\texttt{diagnose} sets a one-shot latch that exactly one subsequent edit consumes. The edit tools refuse unless the latch is set, and \texttt{diagnose}
itself refuses if the rendered views are stale with respect to the
current program. Without the gate the agent edits repeatedly off a stale review; an ungated
variant issued $15$ edits against $2$ diagnoses, whereas the gated loop holds
the diagnose-to-edit ratio at $\approx\!1\!:\!1$.

\needspace{16\baselineskip}%
\begin{wrapfigure}[16]{r}{0.5\textwidth}
  \vspace{-\baselineskip}
  \centering
  \includegraphics[width=0.5\textwidth]{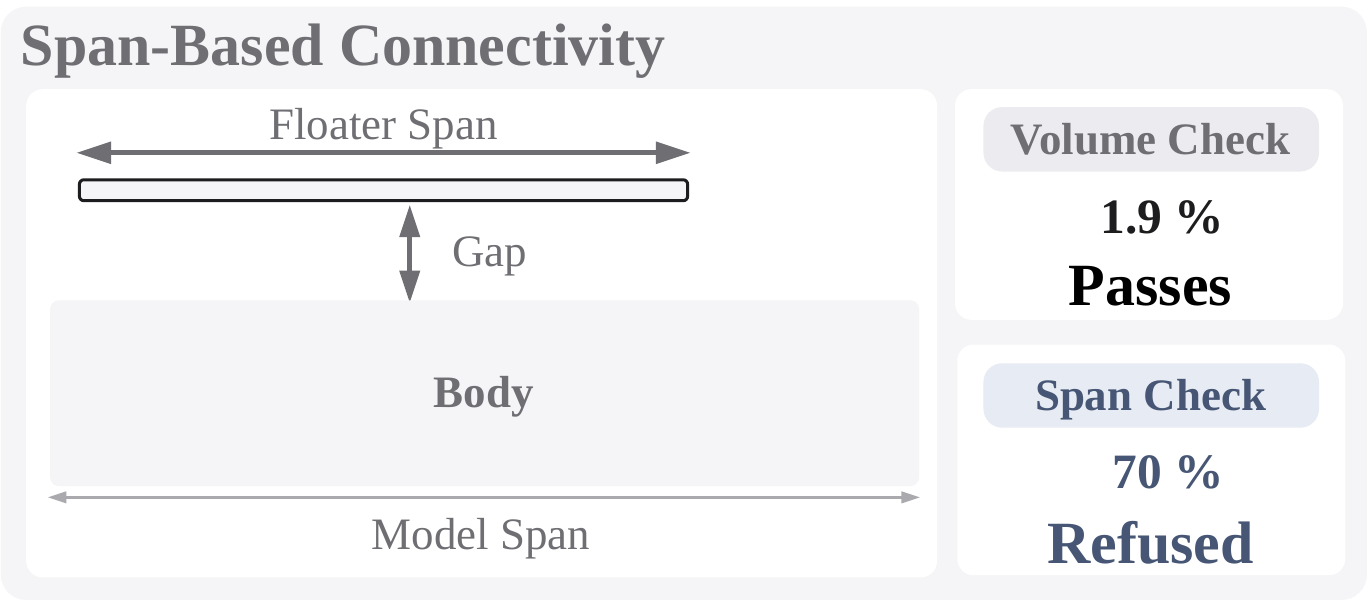}
  \caption{\textbf{Span-based connectivity.} A thin
  detached panel has near-zero enclosed volume, so a volume check passes
  and the floater ships; its bounding-box span covers most of the model
  and the defect is plainly visible. \tether{} thresholds on span
  (\cref{eq:gate}). Drawn to the worst case from our audit of a pre-gate
  pipeline: $70\%$ span, $1.9\%$ volume.}
  \label{fig:span}
\end{wrapfigure}
\runin{Agent-chosen context.}
The agent selects which views to render and how many ($1$ to $20$) from a
fixed catalog of orthographic faces, isometric corners, eye-level diagonals,
and front tilts. On a fresh model it surveys broadly so no side hides a
defect; once a problem is localized it renders the focused subset that best shows
it. A refine step is bounded by a single module edit, and only edits count against the budget. After the edit,
the agent compiles and, when the compile flags floaters, checks connectivity
within the same cycle, so a broken edit is caught before it wastes the next
review. The agent ends with \texttt{ok} when the critic reports no
\texttt{HIGH} issue, or \texttt{give\_up} when several edits fail to shrink
the issue list.

\runin{Connectivity as a hard constraint.}
A faithful-looking render does not guarantee a printable object, since parts
that merely abut compile to several disconnected solids even though they
look joined. The mate gate of \cref{sec:build} prevents most of these by
construction; the refine loop closes the rest with a terminal check on the
mesh that actually ships.

\runin{Span-based gate.}
Let $M$ be the compiled mesh. We weld coincident vertices and partition its
triangles into connected components by union-find over shared edges:
\begin{equation}
  C = \{c_1,\dots,c_K\} = \textsc{components}(M), \qquad
  \rho_i = \frac{\operatorname{span}(c_i)}{\operatorname{span}(M)},
\end{equation}
where $\operatorname{span}(\cdot)$ is the largest side of the axis-aligned
bounding box. Let the body be the largest component, $b = \arg\max_i
\operatorname{vol}(c_i)$. A component $c_i$ with $i \neq b$ is a visible
floater if its span fraction $\rho_i$ reaches a threshold $\tau$, and the
gate passes only when none remain:
\begin{equation}
  N_{\mathrm{vis}} = \bigl|\{\, i \neq b : \rho_i \ge \tau \,\}\bigr|,
  \qquad
  \textsc{gate}(M) = \bigl[\, N_{\mathrm{vis}} = 0 \,\bigr],
  \label{eq:gate}
\end{equation}
with $\tau = 0.01$.

A volume-based criterion instead misses the floaters that occur in practice. A thin, flat panel has near-zero enclosed volume yet a large span, so it passes a volume check while remaining clearly detached (\cref{fig:span}).
We measure by span so the metric tracks what a viewer and the critic actually
see.

\runin{Gated finish.}
The terminal action $\texttt{finish}(\texttt{ok})$ is refused while
$N_{\mathrm{vis}} > 0$. Since the agent may edit after its
last compile, the committed program is recompiled when stale, so the gate
grades the mesh the user actually receives.
When it refuses, it names the worst offenders with their bounding box so the
next cycle has a concrete target. The $\texttt{give\_up}$ verdict is never
gated, allowing a genuinely intentional separation to terminate.

\subsection{Materials}
\label{sec:paint}

\runin{Material library and assignment.}
Geometry alone does not make a production asset, so the same graph also
carries a material for every part. Paint runs once the geometry is frozen and changes only colour and surface; the whole stage is four one-shot vision calls
with no agent loop. \tether{} first reads the reference image and catalogues
every distinct surface as a compact material library, typically one to two
dozen physically-based-rendering (PBR) entries with albedo, class, roughness,
metalness, and wear. It then assigns one library material to each named part,
so the palette is grounded in the reference and shared across parts made of
the same thing. As with geometry, a separate vision critic compares a render of the painted model to the reference and returns a corrected assignment.

\begin{figure}[t]
  \centering
  \includegraphics[width=\textwidth]{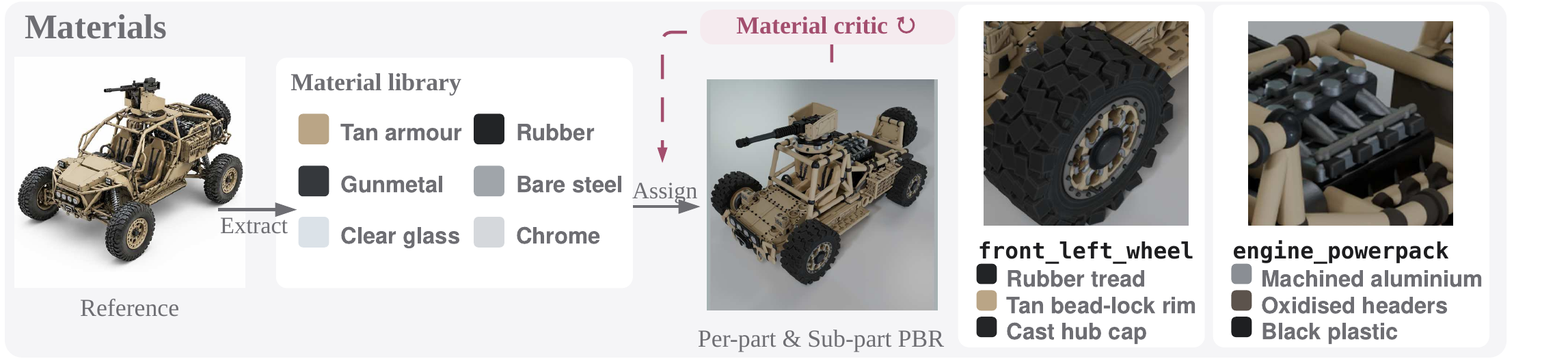}
  \caption{\textbf{Per-part PBR materials from the reference.} \tether{}
  extracts a compact material library from the reference ($25$ entries for the
  assault buggy, six shown), assigns one entry to each of the $41$ named
  parts, and a decoupled material critic corrects the assignment against the
  reference. A sub-part pass then splits a module into sibling colour blocks
  where one material is too coarse. Renders are Blender-Cycles PBR.}
  \label{fig:paint}
\end{figure}

\runin{Sub-part colour blocks.}
One material per module is often too coarse, since one wheel module carries a
rubber tread, a painted rim, and a cast hub. \tether{} therefore rewrites a
module into sibling colour blocks so a sub-feature takes its own material,
recompiling each rewrite and rejecting any whose triangle count moves beyond
a fixed tolerance, so colour can never alter the shape. The stage emits a
colour-tagged program, a multi-group \texttt{OBJ}$+$\texttt{MTL} with
per-part PBR fields, and a Blender-Cycles studio render (\cref{fig:paint});
every artifact stays parts-addressable.

\subsection{Articulation}
\label{sec:motion}

Turning the same parts into a jointed, physically validated asset requires
no new structure, since the mate network is already the kinematic graph. Static mates group parts into rigid links. Each kinematic mate, whether revolute, prismatic, or spherical, names its
joint, axis, and anchor, and is either authored at plan time or promoted
later from geometric evidence.

\runin{Geometric evidence.}
\tether{} measures the geometry before any reasoning about motion. For each part it
computes two kinds of evidence: rotational-symmetry axes, taken from the
area-weighted covariance of the part's surface triangles, and contact
regions between touching parts. Both are deterministic and yield concrete
axis and anchor candidates, so a wheel's spin axis or a hinge line starts
from measured geometry.

\begin{figure}[t]
  \centering
  \includegraphics[width=\textwidth]{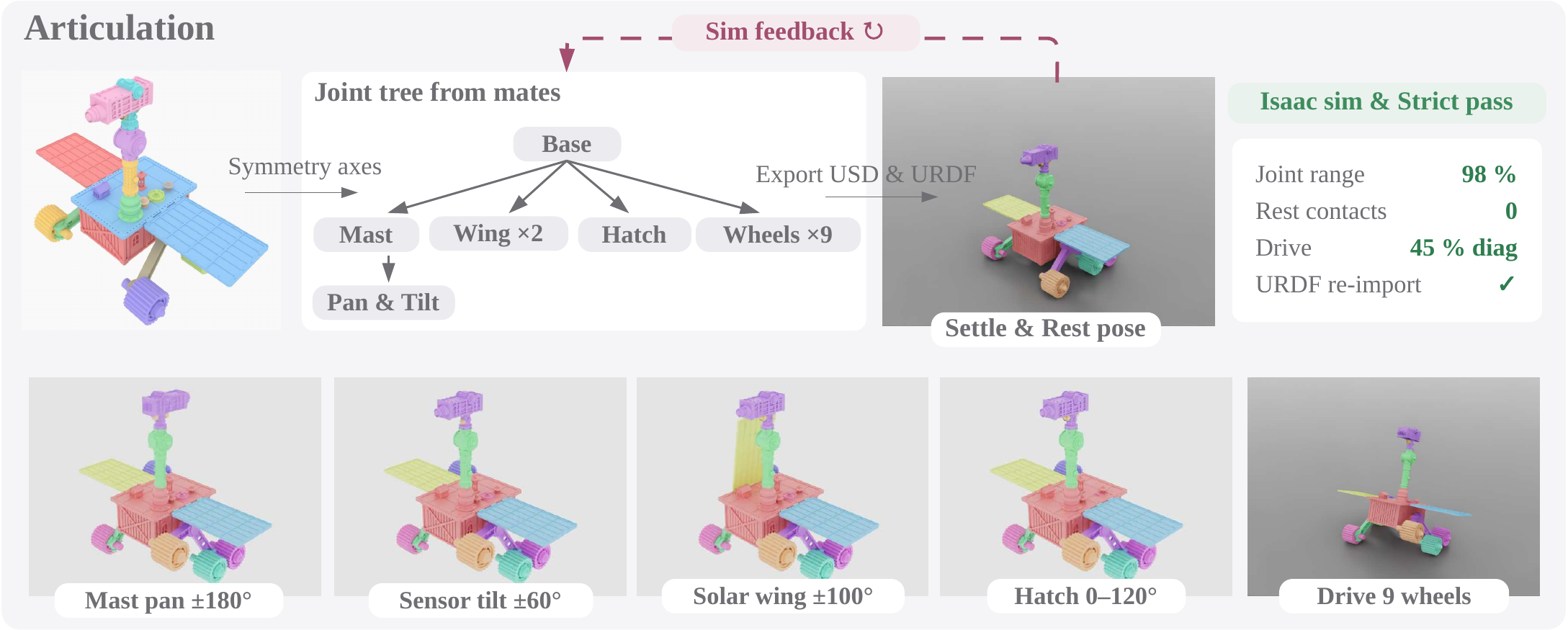}
  \caption{\textbf{A sim-ready articulation validated in Isaac Sim.}
  From the rover's named parts, \tether{} measures geometric evidence (the
  rotational-symmetry axes and contact regions) and plans the joint tree over the kinematic mates ($17$ links, $16$ joints). It exports an OpenUSD
  articulation and a URDF and validates them headless in Isaac Sim; a failed
  verdict feeds rendered simulation frames back to the planner for one refine.
  Every panel is a real Isaac frame in per-link colours: the settled rest pose,
  four per-joint actuation sweeps, and the rover after a nine-wheel drive.}
  \label{fig:motion}
\end{figure}

\runin{Joint planning and validation.}
A vision call then completes the articulation. It groups modules into links, builds the
parent--child tree over the kinematic mates, and assigns each joint its
axes, limits, and drives. The plan is refined against deterministic feedback
and exported as an OpenUSD articulation and a URDF, so the same object drops
into a physics simulator or a robotics stack. \tether{} then validates the
plan headless in Isaac Sim through a fixed battery: a dynamic settle,
per-joint actuation sweeps, a rest-contact scan, a wheeled mobility test,
and a URDF re-import. A flagged fault feeds up to eight rendered simulation
frames back to the planner for one refine, so the critic can see a wrong
axis
(\cref{fig:motion}).

\section{Experiments}
\label{sec:exp}

Our experiments answer three questions. Does mate-driven shape-as-code
make a frozen LLM a state-of-the-art 3D generator, measured against native
generators, prior 3D-code agents, and single-shot prompting of the same
models (\cref{sec:cross,sec:p3d})? What do the pipeline's stages contribute
(\cref{sec:ablation})? And does the program deliver what a mesh
cannot, namely a full production asset (\cref{sec:extras})? Throughout we report intervals and paired
tests rather than single numbers.

\subsection{Setup and Protocol}
\label{sec:setup}

\runin{Benchmarks.}
We evaluate on two benchmarks. The first is MechBench-36, our hard-surface
benchmark, a frozen set of $36$ deliberately many-part objects, rovers, mechs, cranes,
excavators, engines, and landing gear, chosen because they stress part
counts, placement, and connectivity far harder than a single-sub-assembly
shape (\cref{sec:cross}). The second is the assembly task of
P3D-Bench~\cite{p3dbench}, $203$ text-and-image cases scored by the
benchmark's own assembly judge (\cref{sec:p3d}). \tether{} normally synthesizes its own reference
view from the prompt; for evaluation we lock both the prompt and that view
and hand the same pair to every method, so image synthesis cannot confound
the comparison.

\runin{Blind judging protocol.}
Judging generated 3D by eye invites bias, so the protocol removes each
confound in turn. Every shipped model is re-rendered from one fixed view set
under one grey-clay rig, assigned an opaque hash, and judged blind, so method
names never reach the judge and render settings cannot skew a row. A vision
LLM~\cite{wu2024gpteval3d} scores each render with one shared prompt on three
axes, semantic fidelity, geometric quality, and aesthetics, the last rating
the object as an engineered part. On MechBench-36 the judge sees no ground truth, since none exists for a prompt; on P3D-Bench it follows
the benchmark's own protocol. Alongside the judge we report CLIP, the image-text alignment between the
prompt and the shared grey-clay renders, and two edge measurements on the
shipped mesh, sharp-edge length and the $95$th-percentile dihedral
(\cref{tab:hs36}). All are computed under the one shared rig, so they are comparable across rows but not across papers. 

\runin{Models and implementation.}
Unless noted, planning, building, and the critic all use one frozen
deep-reasoning LLM, Gemini 3.7 Flash, with no 3D-specific training;
\cref{tab:p3d} additionally reports a GPT-5.6-sol variant of the full
pipeline. The judge is Gemini 3.7 Flash in its high-reasoning mode, blinded
by hash. \tether{} is built on an in-house LLM harness; the program is realized
as OpenSCAD and compiled by the OpenSCAD engine, with module parsing, splicing,
the mesh parser, and the union-find connectivity analysis in process. Beyond
the compiler, the only external binaries are Blender for renders and a headless
Isaac Sim for physical validation.

\subsection{Results on MechBench-36}
\label{sec:cross}

\begin{table}[t]
  \centering\footnotesize
  \setlength{\tabcolsep}{4pt}
  \begin{tabular}{@{}lcccccrr@{}}
    \toprule
    & \multicolumn{4}{c}{Judge} & & \multicolumn{2}{c}{Edge structure} \\
    \cmidrule(lr){2-5}\cmidrule(lr){7-8}
    Method & Geo$\uparrow$ & Aes$\uparrow$ & Sem$\uparrow$ & Overall$\uparrow$ & CLIP$\uparrow$ & Shrp$_{60}\uparrow$ & Dih$_{95}\uparrow$ \\
    \midrule
    \textbf{\tether{} (Gemini 3.7 Flash)} & \textbf{0.799} & \textbf{0.827} & 0.858 & \textbf{0.828} & \textbf{82.67} & 158.08 & 97.97 \\
    \textbf{\tether{} (GPT-5.6-sol)} & 0.756 & 0.793 & 0.843 & 0.797 & 81.76 & \textbf{185.18} & \textbf{105.28} \\
    \midrule
    \multicolumn{8}{@{}l}{\emph{Agentic 3D-code}} \\
    Adam CAD (GPT-5.6-sol)~\cite{adamcad}            & 0.778 & 0.759 & \textbf{0.861} & 0.799 & 81.26 &  57.35 & 90.41 \\
    ArtiCraft (GPT-5.6-sol)~\cite{zhou2026articraft} & 0.682 & 0.596 & 0.827 & 0.702 & 79.12 &  41.31 & 94.74 \\
    CAD-Coder~\cite{guan2025cadcoder}            & 0.099 & 0.210 & 0.096 & 0.135 & 60.62 &   7.63 & 92.91 \\
    cadrille~\cite{kolodiazhnyi2026cadrille}     & 0.086 & 0.105 & 0.083 & 0.092 & 50.65 &  19.58 & 97.18 \\
    \midrule
    \multicolumn{8}{@{}l}{\emph{Single-shot LLM}} \\
    GPT-5.6-sol      & 0.790 & 0.735 & 0.852 & 0.792 & 80.91 &  60.39 & 89.76 \\
    Gemini 3.7 Flash & 0.719 & 0.722 & 0.849 & 0.763 & 80.86 &  63.54 & 91.35 \\
    Gemini 3.1 Pro   & 0.698 & 0.673 & 0.824 & 0.731 & 79.94 &  --- & --- \\
    \midrule
    \multicolumn{8}{@{}l}{\emph{Native 3D generation}} \\
    TRELLIS.2~\cite{xiang2025trellis2}   & 0.781 & 0.787 & \textbf{0.861} & 0.810 & 77.51 & 134.02 & 73.11 \\
    UltraShape~\cite{jia2025ultrashape} & 0.731 & 0.694 & 0.818 & 0.748 & 75.04 &   8.58 & 33.06 \\
    Hunyuan3D~\cite{hunyuan3d2025}      & 0.685 & 0.590 & 0.815 & 0.697 & 74.62 &   6.41 & 35.12 \\
    Direct3D-S2~\cite{wu2025direct3ds2} & 0.583 & 0.559 & 0.775 & 0.639 & 74.90 &  35.17 & 52.78 \\
    \bottomrule
  \end{tabular}
  \caption{\textbf{MechBench-36} (\cref{fig:cmp-llm,fig:cmp-native}). Every method is re-rendered through one rig and scored by the same judge
  and CLIP backbone, so the only difference between rows is the method. Geo, Aes, and Sem are the judge's geometry, aesthetics, and semantic axes,
  mapped per case to $0$--$1$ by $\mathrm{clamp}((v-1)/9)$ as in \cref{tab:p3d};
  aesthetics rates the object as an engineered part, and Overall is their
  mean. CLIP is prompt-to-render alignment on the shared renders. Shrp$_{60}$ is total sharp-edge length above a
  $60^\circ$ dihedral and Dih$_{95}$ the $95$th-percentile dihedral, both
  measured on the shipped mesh. The $60^\circ$ threshold keeps only genuine machined creases, since a
  shallower one also counts the ridges of a regressed surface. Cases where a method shipped no mesh score zero and
  stay in the mean, which is why CAD-Coder and cadrille sit low. Single-shot rows are one call with no plan, build, or refine stage. Best per
  column in bold.}
  \label{tab:hs36}
\end{table}

MechBench-36 concentrates the many-part regime that motivates
\tether{}. As shown in \cref{tab:hs36},
\tether{} takes the top judge composite at $0.828$ with Gemini 3.7 Flash,
leads geometry, aesthetics, and CLIP, and ships a mesh for every case.
TRELLIS.2 follows at $0.810$; Adam CAD at $0.799$ and our GPT-5.6-sol arm at
$0.797$ are effectively tied.

\runin{Axis-level analysis.}
The judge's axes do not separate the field equally. Semantic spans only $0.775$
to $0.861$ across the methods that work at all, since that axis mostly asks
whether the object is of the right kind; Adam CAD and TRELLIS.2 share its top
at $0.861$, a whisker above our $0.858$. Geometry and aesthetics separate the
methods far more, spanning $0.583$ to $0.799$ and $0.105$ to $0.827$, and
\tether{} tops both.

\begin{figure}[p]
  \centering
  \captionsetup{skip=4pt}
  \resizebox{0.92\linewidth}{!}{\begin{minipage}{\linewidth}\centering
  \makebox[0.97\linewidth][l]{%
    \rlap{\hspace*{0.092\linewidth}\makebox[0pt][c]{\scriptsize \textbf{\tether{} (ours)}}}%
    \rlap{\hspace*{0.291\linewidth}\makebox[0pt][c]{\scriptsize TRELLIS.2}}%
    \rlap{\hspace*{0.490\linewidth}\makebox[0pt][c]{\scriptsize Hunyuan3D-2.1}}%
    \rlap{\hspace*{0.690\linewidth}\makebox[0pt][c]{\scriptsize UltraShape}}%
    \rlap{\hspace*{0.886\linewidth}\makebox[0pt][c]{\scriptsize Direct3D-S2}}%
  }\par\vspace{0.4pt}
  \includegraphics[width=0.97\linewidth]{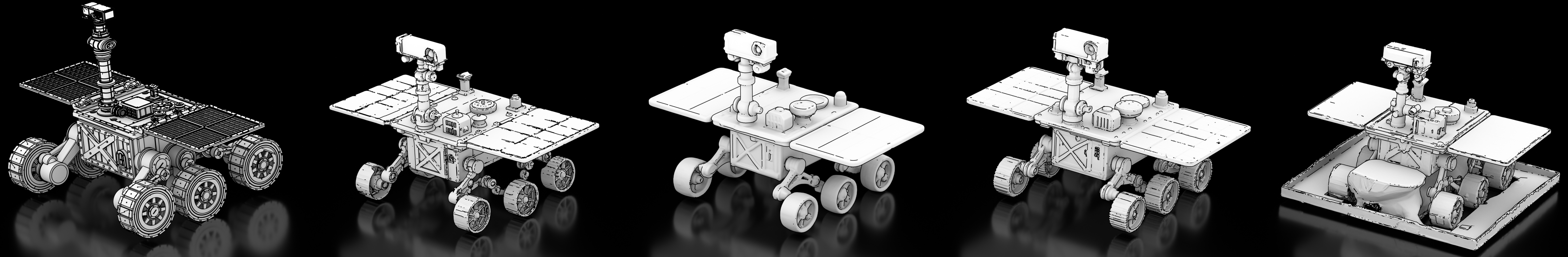}\par
  {\scriptsize\itshape A six-wheeled Mars rover with a boxy chassis, an articulated camera mast, and broad flat solar-panel wings.}\par\vspace{0.4pt}
  \includegraphics[width=0.97\linewidth]{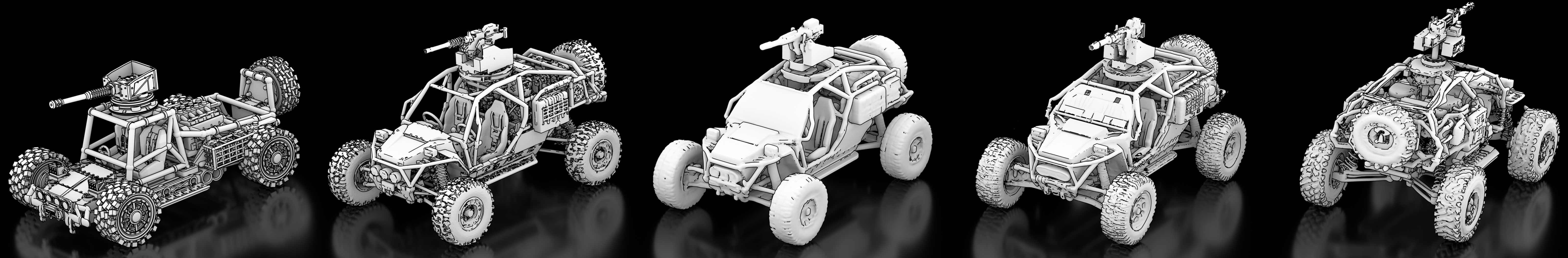}\par
  {\scriptsize\itshape An armed assault buggy with a tubular roll cage, long-travel suspension, a roof turret, and a rear spare wheel.}\par\vspace{0.4pt}
  \includegraphics[width=0.97\linewidth]{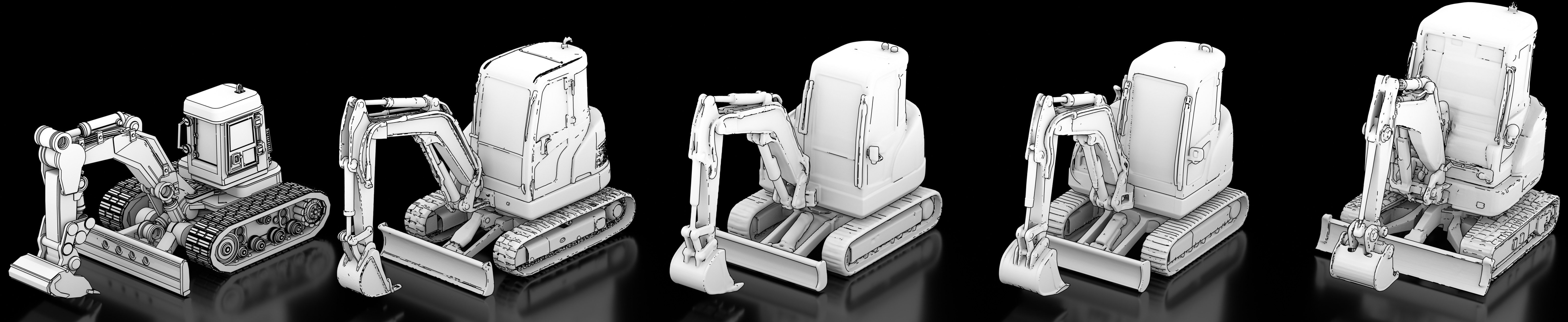}\par
  {\scriptsize\itshape A compact tracked excavator with cab, boom, stick, bucket, hydraulic cylinders, and a dozer blade.}\par\vspace{0.4pt}
  \includegraphics[width=0.97\linewidth]{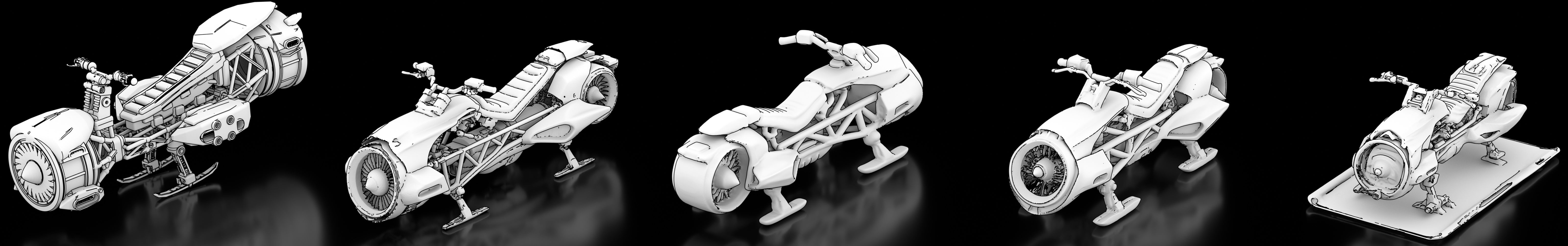}\par
  {\scriptsize\itshape A hover speeder bike with a sculpted spine seat, fore and aft ducted turbines, side fairings, and landing skids.}\par\vspace{0.4pt}
  \includegraphics[width=0.97\linewidth]{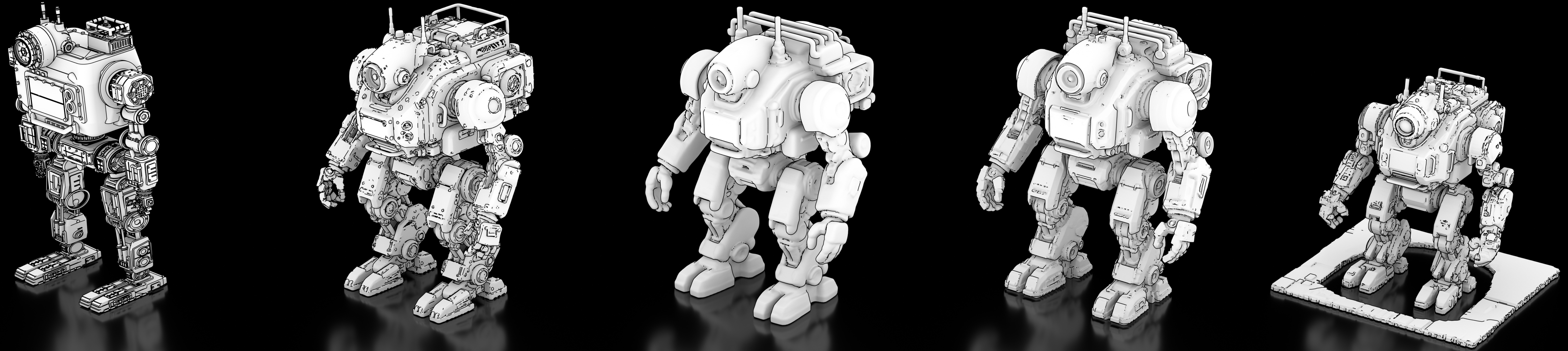}\par
  {\scriptsize\itshape A retro-futuristic bipedal exploration mech with armored panels, piston legs, and a visored sensor head.}\par\vspace{0.4pt}
  \includegraphics[width=0.97\linewidth]{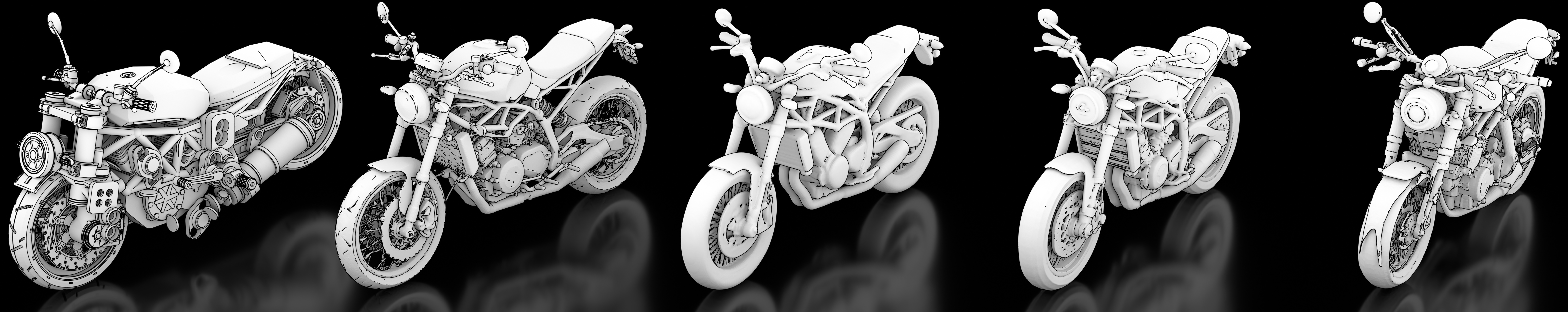}\par
  {\scriptsize\itshape A motorcycle rolling chassis with trellis frame, V-twin engine, fork, swingarm, spoked wheels, and exhaust.}
  \end{minipage}}
  \caption{\textbf{Qualitative comparison against native 3D generators.} The same six objects, one per row with its prompt printed beneath it, generated by four native 3D diffusion models and by \tether{} (left column), all conditioned on the same locked reference image and re-rendered from the same camera and shading rig. The natives reproduce the silhouette faithfully, since they are trained end-to-end to do so, but their surfaces are soft, so bolt heads, vents, and panel lines fuse into one skin and no part boundary survives. \tether{} compiles exact CSG solids, so edges stay sharp and every part remains separable.}
  \label{fig:cmp-native}
\end{figure}

\runin{Comparison with native generators.}
Native generators map an image to a dense surface whose creases come out rounded. The strongest of them, TRELLIS.2,
trails by $5.2$ CLIP points, and Direct3D-S2 and Hunyuan3D fall behind on
aesthetics by $0.27$ and $0.24$. In \cref{fig:cmp-native}, the natives reproduce the silhouette faithfully but blur bolt heads, vents, and panel lines into the surrounding body, whereas our compiled CSG keeps every feature crisp and every part separable.

\begin{figure}[p]
  \centering
  \captionsetup{skip=4pt}
  \resizebox{0.92\linewidth}{!}{\begin{minipage}{\linewidth}\centering
  \makebox[0.97\linewidth][l]{%
    \rlap{\hspace*{0.095\linewidth}\makebox[0pt][c]{\scriptsize \textbf{\tether{} (ours)}}}%
    \rlap{\hspace*{0.292\linewidth}\makebox[0pt][c]{\scriptsize GPT-5.6 (img)}}%
    \rlap{\hspace*{0.491\linewidth}\makebox[0pt][c]{\scriptsize Gemini 3.1 Pro (img)}}%
    \rlap{\hspace*{0.686\linewidth}\makebox[0pt][c]{\scriptsize Adam CAD}}%
    \rlap{\hspace*{0.887\linewidth}\makebox[0pt][c]{\scriptsize ArtiCraft}}%
  }\par\vspace{0.4pt}
  \includegraphics[width=0.97\linewidth]{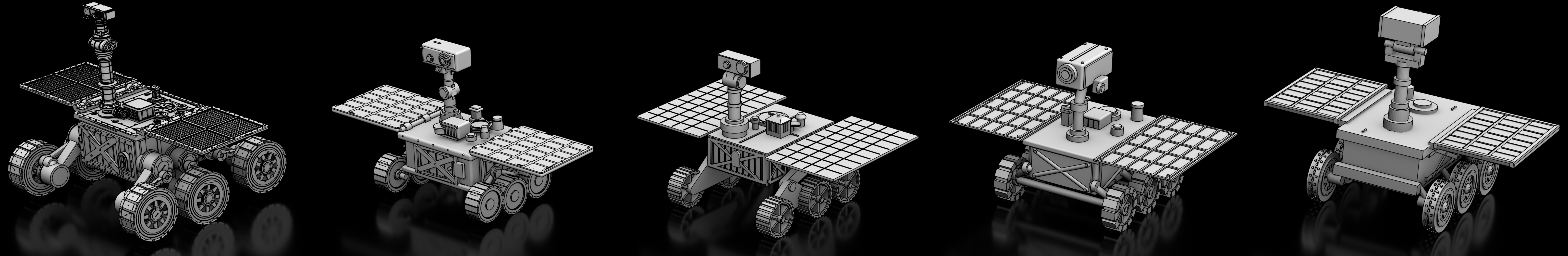}\par
  {\scriptsize\itshape A six-wheeled Mars rover with a boxy chassis, an articulated camera mast, and broad flat solar-panel wings.}\par\vspace{0.4pt}
  \includegraphics[width=0.97\linewidth]{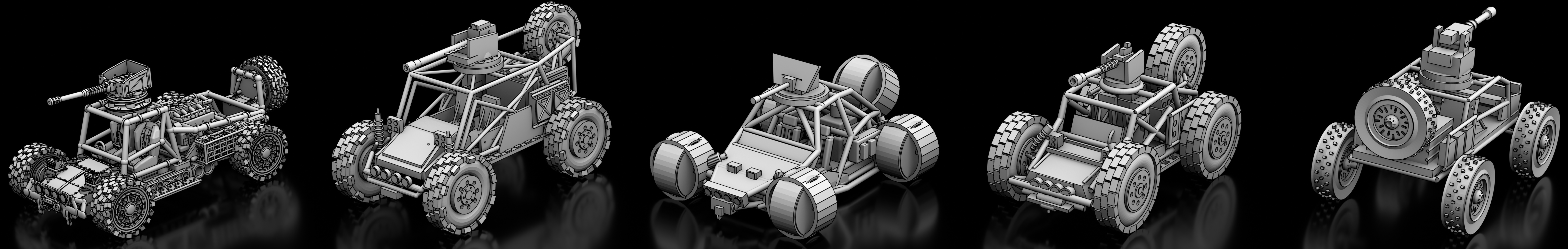}\par
  {\scriptsize\itshape An armed assault buggy with a tubular roll cage, long-travel suspension, a roof turret, and a rear spare wheel.}\par\vspace{0.4pt}
  \includegraphics[width=0.97\linewidth]{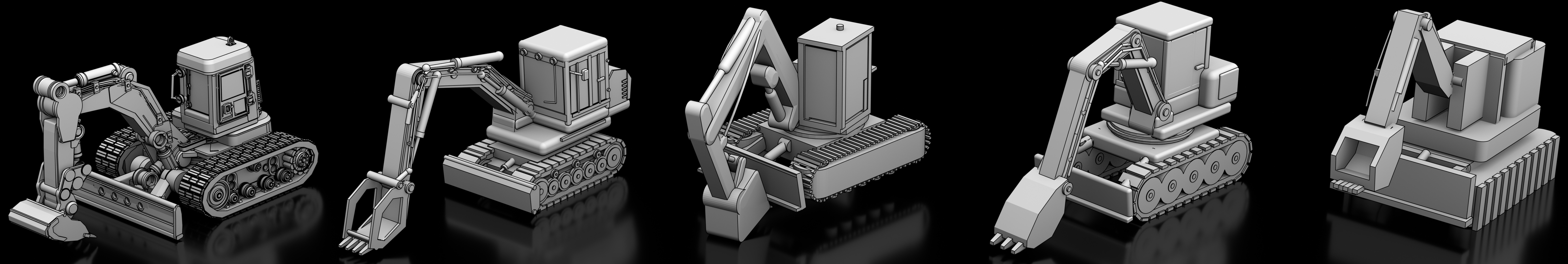}\par
  {\scriptsize\itshape A compact tracked excavator with cab, boom, stick, bucket, hydraulic cylinders, and a dozer blade.}\par\vspace{0.4pt}
  \includegraphics[width=0.97\linewidth]{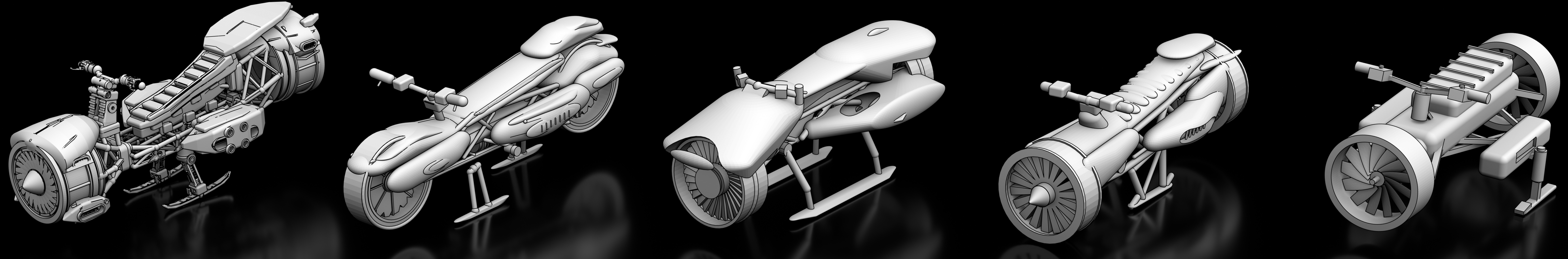}\par
  {\scriptsize\itshape A hover speeder bike with a sculpted spine seat, fore and aft ducted turbines, side fairings, and landing skids.}\par\vspace{0.4pt}
  \includegraphics[width=0.97\linewidth]{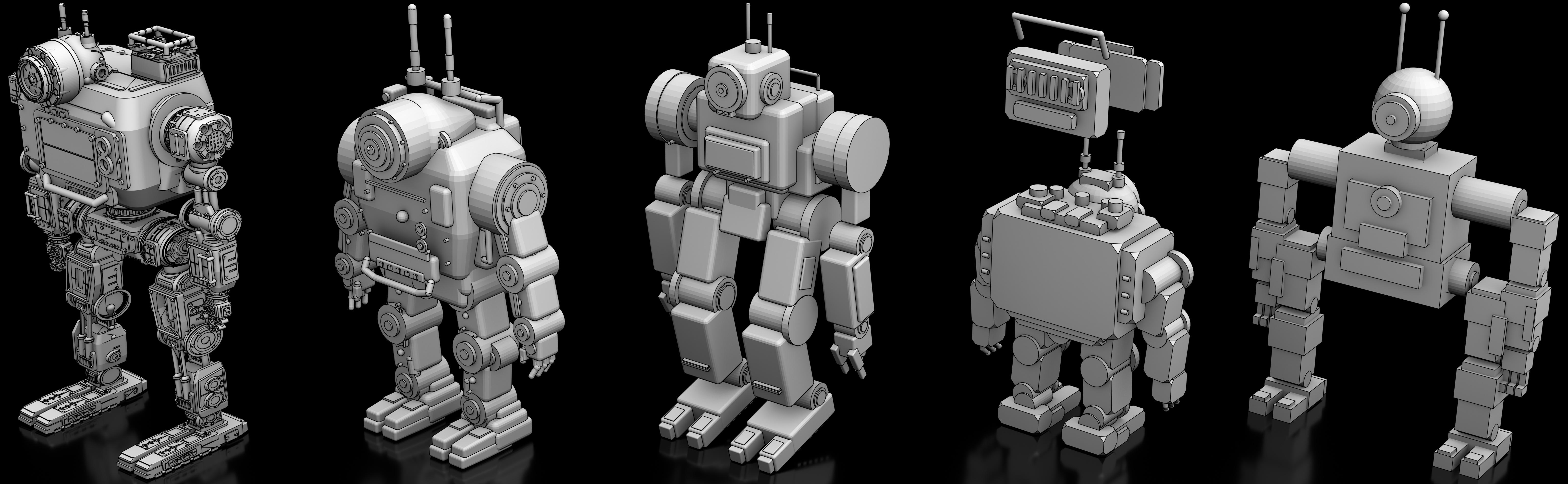}\par
  {\scriptsize\itshape A retro-futuristic bipedal exploration mech with armored panels, piston legs, and a visored sensor head.}\par\vspace{0.4pt}
  \includegraphics[width=0.97\linewidth]{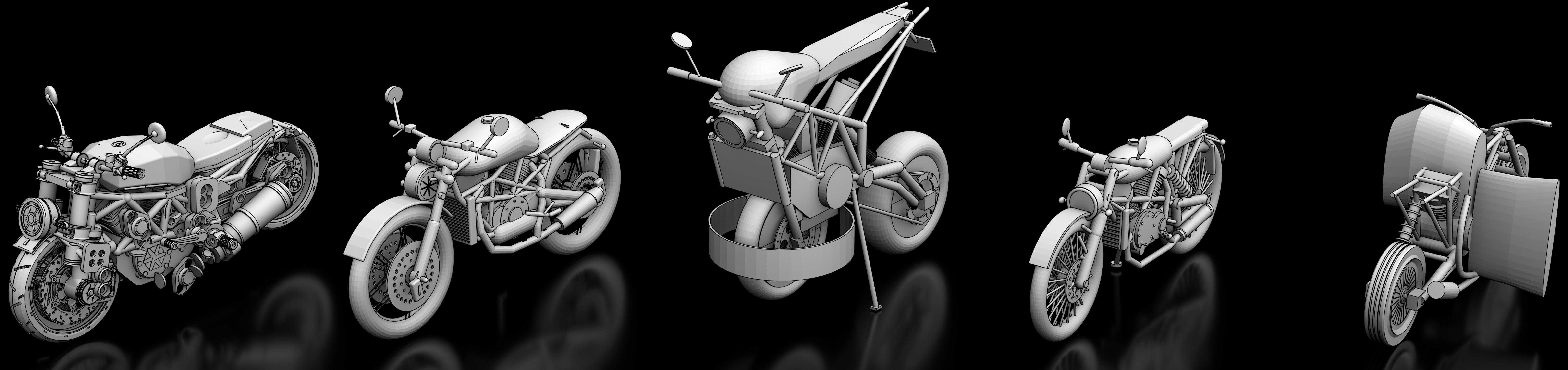}\par
  {\scriptsize\itshape A motorcycle rolling chassis with trellis frame, V-twin engine, fork, swingarm, spoked wheels, and exhaust.}
  \end{minipage}}
  \caption{\textbf{Qualitative comparison against LLM-driven 3D generation.} Six
  benchmark objects, one per row, each generated from the prompt printed
  beneath it; every method receives the same reference image and render rig.
  Prior agents recover the rough silhouette but collapse the mechanism into a
  handful of boxes, whereas \tether{} resolves each part as its own solid.}
  \label{fig:cmp-llm}
\end{figure}

\runin{Comparison with prior 3D-code agents.}
The strongest prior agent, Adam CAD on GPT-5.6-sol, reaches $0.759$
aesthetics and $81.26$ CLIP. The domain-tuned coders CAD-Coder and cadrille collapse on
many-part objects, scoring $0.210$ and $0.105$ and shipping no mesh on $1$ and $5$ cases, since a model trained to emit one sketch-extrude body cannot
express an assembly. The same failure is visible in \cref{fig:cmp-llm}, where the prior agents keep the outline of each machine but simplify its mechanism into a few large blocks.

\runin{Edge structure.}
The two right-hand columns measure the edge structure each representation
produces, and \tether{} leads both. Our programs carry $158.1$ units of sharp edge above a $60^\circ$ dihedral
with Gemini 3.7 Flash and $185.2$ with GPT-5.6-sol, $2.6\times$ and
$3.1\times$ the next code-emitting method, at $95$th-percentile dihedrals of
$98.0^\circ$ and $105.3^\circ$. The native
generators sit at $33$--$73^\circ$, since a surface regressed from samples
has no notion of a face or a bore. TRELLIS.2 posts the only comparable
sharp-edge length at $134.0$, yet its $73.1^\circ$ dihedral shows those
creases are far shallower than ours. The gap follows from the representation, since an edge in compiled CSG is
the exact intersection of two analytic surfaces.

\runin{Paired gain over the base model.} Every row uses a frozen model, so the pipeline can be ablated against a
single call to the same model on the same case. Paired per object over the
whole benchmark, the pipeline improves aesthetics by $+0.105$ on Gemini 3.7
Flash ($20$ wins, $7$ losses, sign test $p = 0.019$); on GPT-5.6-sol the mean
gain is $+0.059$ but the split is $15$ wins to $11$ losses ($p = 0.56$), so
we do not claim significance there. CLIP improves by $+1.80$ ($p = 0.011$)
and $+0.85$ respectively. The gain concentrates on aesthetics, the axis the mate-driven build targets.

\subsection{Results on P3D-Bench}
\label{sec:p3d}

\begin{table}[t]
  \centering\small
  \setlength{\tabcolsep}{6pt}
  \begin{tabular}{@{}lcccc@{}}
    \toprule
    Method & Semantic$\uparrow$ & Geometry$\uparrow$ & Aesthetics$\uparrow$ & Judge (Overall)$\uparrow$ \\
    \midrule
    \textbf{\tether{} (Gemini 3.7 Flash)} & \textbf{0.666} & \textbf{0.490} & \textbf{0.616} & \textbf{0.590}\,{\scriptsize$\pm$0.009} \\
    \tether{} (GPT-5.6-sol) & 0.661 & 0.460 & 0.604 & 0.575\,{\scriptsize$\pm$0.010} \\
    \midrule
    Gemini 3.7 Flash & 0.662 & 0.458 & 0.577 & 0.566\,{\scriptsize$\pm$0.016} \\
    GPT-5.6-sol      & 0.662 & 0.444 & 0.584 & 0.563\,{\scriptsize$\pm$0.015} \\
    GPT-5.5      & 0.643 & 0.392 & 0.537 & 0.524\,{\scriptsize$\pm$0.012} \\
    Gemini 3.1 Pro & 0.639 & 0.390 & 0.508 & 0.513\,{\scriptsize$\pm$0.012} \\
    Opus 4.6     & 0.572 & 0.327 & 0.446 & 0.448\,{\scriptsize$\pm$0.016} \\
    Kimi K2.6    & 0.523 & 0.294 & 0.372 & 0.396\,{\scriptsize$\pm$0.014} \\
    Qwen 3.6     & 0.451 & 0.264 & 0.322 & 0.346\,{\scriptsize$\pm$0.014} \\
    Doubao       & 0.408 & 0.253 & 0.302 & 0.321\,{\scriptsize$\pm$0.014} \\
    GLM-5V Turbo & 0.402 & 0.249 & 0.300 & 0.317\,{\scriptsize$\pm$0.013} \\
    MiMo Omni    & 0.379 & 0.242 & 0.294 & 0.305\,{\scriptsize$\pm$0.012} \\
    \bottomrule
  \end{tabular}
  \caption{\textbf{P3D-Bench assembly results.} Every method is re-rendered
  from the same rig and scored by the benchmark's assembly judge on its three
  axes, semantic fidelity, geometric quality, and aesthetics, normalized to
  $0$--$1$ by the benchmark's mapping; Judge (Overall) is their mean, the
  benchmark's composite, with $95\%$ intervals. Baselines are the benchmark's
  entries, re-run under the same harness, plus single-shot prompting of the two
  base models \tether{} uses.}
  \label{tab:p3d}
\end{table}

P3D-Bench's judge scores each case on the three axes of \cref{sec:setup},
and its composite is the mean of the three after normalization. We re-render and re-judge every method under the shared
harness, so every row of \cref{tab:p3d} is directly comparable.

As shown in \cref{tab:p3d}, \tether{} solves all $203$ cases with either
base model and leads every column. The composite reaches $0.590$ on Gemini
3.7 Flash and $0.575$ on GPT-5.6-sol, against $0.524$ for the strongest entry
that also covers the full benchmark, a margin several times the $95\%$
intervals. Notably, the lead is largest on geometry, $0.490$ against $0.392$,
which is exactly where the mate-driven build acts, since solved placement
and verified connectivity are geometric properties.

\runin{Gain over single-shot prompting.} Prompting our two base models once reaches $0.566$ and $0.563$ on the reduced
sets those runs cover, above every published entry, so one call to a frontier model already clears the published leaderboard. Part of any margin over the published rows therefore reflects base-model
strength rather than method; the agent supplies the remaining margin. A per-case paired comparison puts the full pipeline above the Gemini 3.7
Flash control on the cases it covers, and the pipeline also solves the cases
the control does not.

\subsection{Ablation Study}
\label{sec:ablation}

\begin{table}[t]
  \centering\small
  \setlength{\tabcolsep}{6pt}
  \begin{tabular}{@{}lccccc@{}}
    \toprule
    Variant & Geo$\uparrow$ & Aes$\uparrow$ & Sem$\uparrow$ & Judge$\uparrow$ & CLIP$\uparrow$ \\
    \midrule
    \tether{} (full) & 0.799 & 0.827 & 0.858 & 0.828 & 82.67 \\
    w/o refine       & 0.762 & 0.793 & 0.846 & 0.800 & 81.84 \\
    w/o planning     & 0.756 & 0.775 & 0.843 & 0.791 & 82.36 \\
    w/o 3D feedback  & 0.790 & 0.815 & 0.852 & 0.819 & 81.96 \\
    \bottomrule
  \end{tabular}
  \caption{\textbf{Ablations on MechBench-36.} All variants use the same base
  model and are judged in the same single session as \cref{tab:hs36}, with the same
  normalization ($n=36$). Ablated variants are evaluated at the draft stage, before
  refine, so the planning and 3D-feedback deltas also include the absence of
  the refine stage.}
  \label{tab:ablation}
\end{table}

\cref{tab:ablation} removes the refine loop, the assembly planning, or the
per-part 3D feedback and re-judges the $36$ cases under the same regime as
\cref{tab:hs36}. Every removal costs quality on every column, and the full
pipeline is the best variant on all five. Dropping the refine loop lowers the
composite from $0.828$ to $0.800$, the cleanest read of what the decoupled
critic's gated edits add, since that variant differs from the full pipeline
by refine alone. Dropping the planning stage costs the most, $0.791$ overall
with the largest aesthetics drop, so the assembly graph the build consumes is
load-bearing. Dropping the per-part 3D feedback lowers every judge axis and
CLIP as well. The deficit grows with how much of the design is removed,
consistent with each stage contributing to the final quality.

\begin{figure}[t]
  \centering
  \captionsetup{skip=4pt}
  \includegraphics[width=0.85\textwidth]{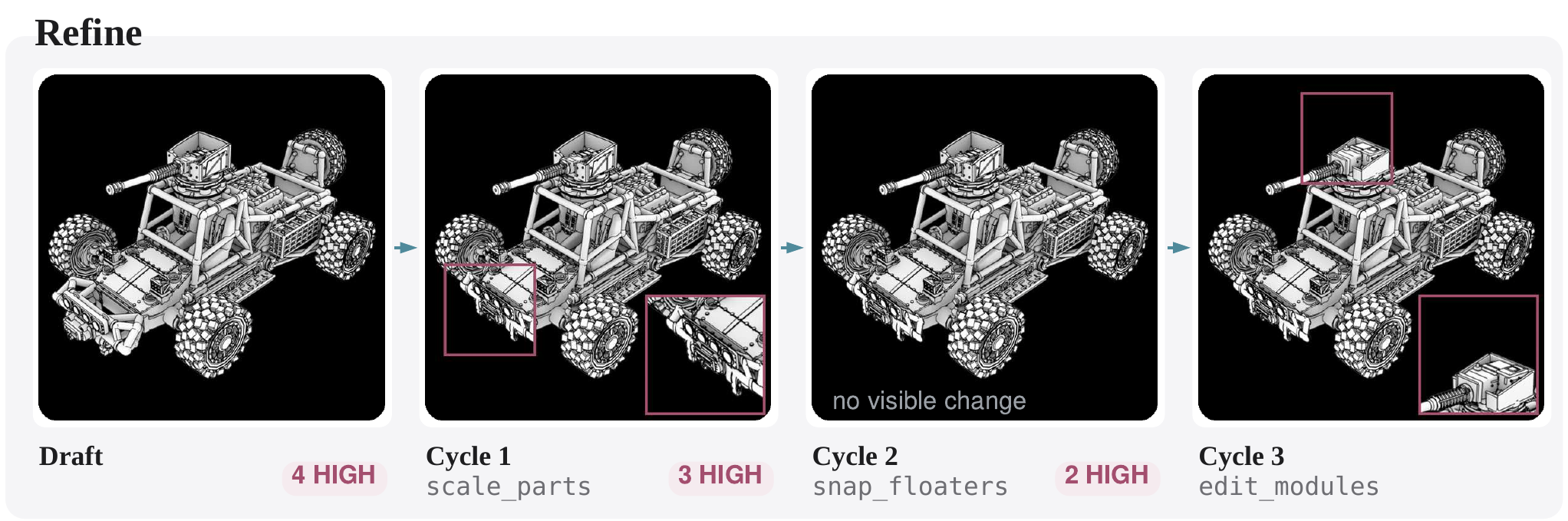}
  \caption{\textbf{What each gated edit changes.} The assault buggy, refined
  from the draft the build stage produced. Every frame is rendered from one
  fixed camera, so a difference between frames is a difference in geometry; the inset magnifies the densest region of pixels that cycle's edit
  changed, and the outline on
  the model marks where the inset is taken from. Each cycle is one fresh
  diagnosis followed by one gated edit; under every state we print the tool
  that produced it and the number of \texttt{HIGH} issues the critic reported
  for that state. The count falls $4\!\rightarrow\!3\!\rightarrow\!2$; the
  final state carries no count because the cycle budget ends before a further
  diagnosis. Cycle~2 is a sub-millimetre contact snap that changes connectivity
  but nothing a reader could see, and the frame says so.}
  \label{fig:refineabl}
\end{figure}

\runin{Refinement case study.}
\cref{fig:refineabl} traces the refine loop on one of the hardest benchmark
cases. The critic's complaints are quantitative and structural. It reports the
front winch $1.25\times$ a road-tire diameter ahead of the front axle against
roughly $0.6$--$0.7\times$ in the reference, and a gun receiver at $0.19\times$
the tire diameter against $0.30$--$0.35\times$, and the agent answers each
with a single tool call on the named modules. Over three cycles the
\texttt{HIGH} list shrinks monotonically, and the history-aware critic carries
context forward, with cycle~3 opening on the rear equipment stack that
``remains too low''. The case ends on \texttt{max-steps} rather than
converging, since one gated fix per fresh diagnosis makes progress linear in
the cycle budget. One of the three edits is a \texttt{snap\_floaters} call
that closes sub-millimetre gaps, changing connectivity without changing the
silhouette, which is why the cycle-2 frame looks unchanged and why we report
the issue list rather than a pixel difference as the outcome.

\begin{figure}[!t]
  \centering
  \captionsetup{skip=4pt}
  \includegraphics[width=0.92\textwidth]{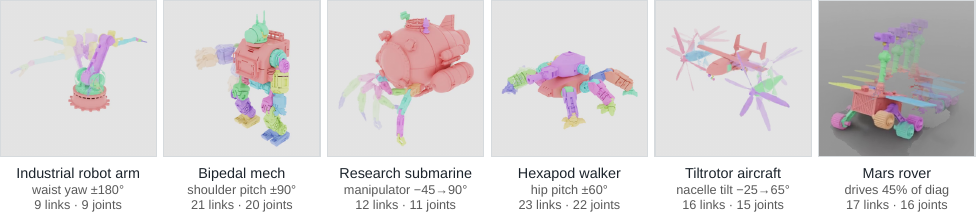}
  \caption{\textbf{Articulation across the benchmark.} Six objects, each
  planned, exported to OpenUSD and URDF, and validated headless in Isaac Sim.
  Every panel is composited from that object's own simulation clip, with the
  swept poses ghosted and the final pose solid, so a still carries the motion.
  The first five hold one commanded joint traversing its range; the last shows
  the wheeled base driving on the ground plane. All six pass every dynamics
  phase: each position-driven joint traverses $\ge\!97\%$ of its commanded
  range, no non-adjacent links touch at rest, and the URDF re-imports. Five
  also clear the asset rules outright; the tiltrotor trips the advisory
  drive-gain rule, the same rule two earlier rover plans tripped.}
  \label{fig:articulation}
\end{figure}

\subsection{Materials and Articulation}
\label{sec:extras}

\begin{figure}[t]
  \centering
  \captionsetup{skip=4pt}
  \includegraphics[width=\textwidth]{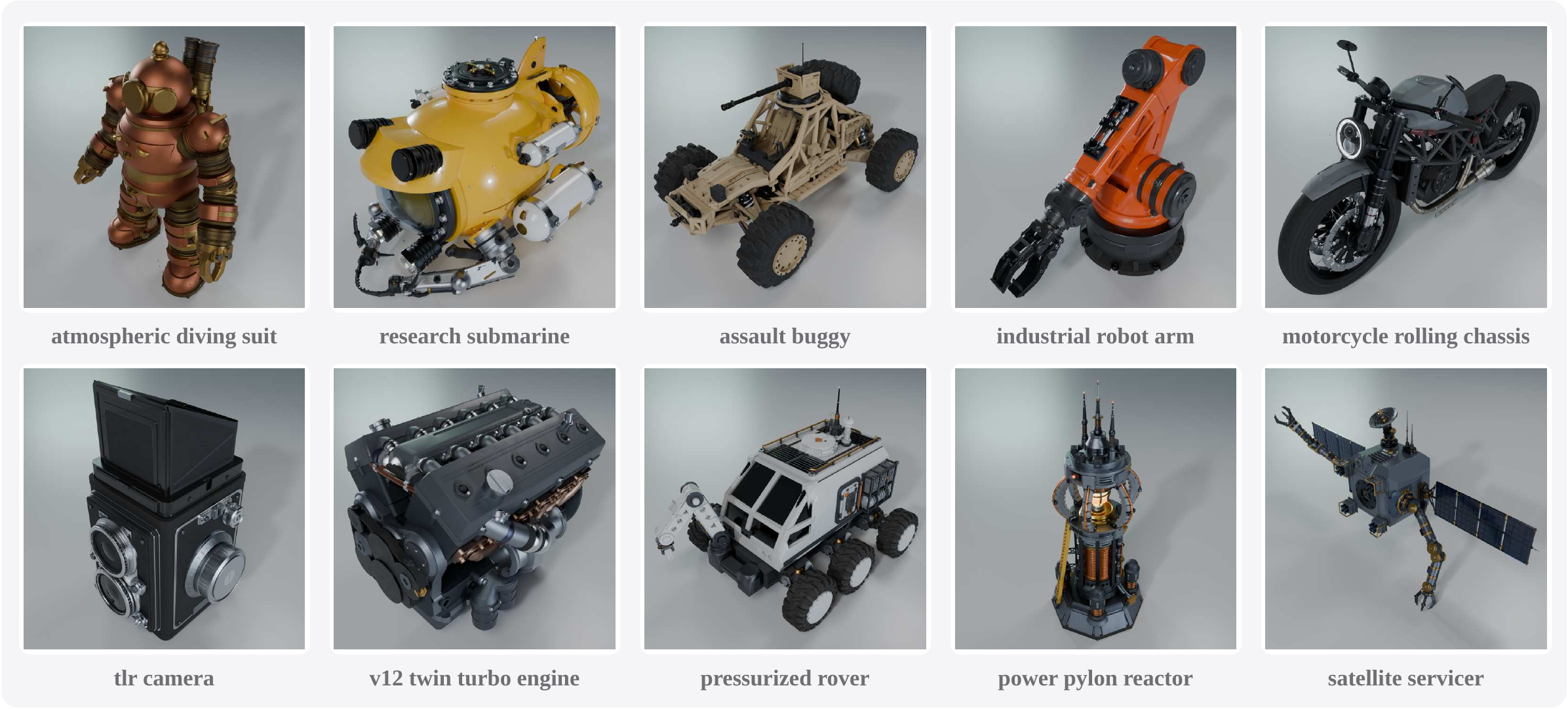}
  \caption{\textbf{Texturing across the benchmark.} Ten of the $36$ benchmark
  objects, geometry and materials both authored by \tether{}. Every image is a
  Blender-Cycles render of the shipped program, with per-part materials
  assigned to the named modules.}
  \label{fig:deeppaint}
\end{figure}

\runin{Materials.}
Because every part is a named module, a physically-based material is a
per-part attribute the program simply carries. \cref{fig:paint} shows an
object painted from its reference alone, where the extract, assign, critic,
and sub-part passes recover distinct metals, rubber, and glass, and
\cref{fig:deeppaint} shows the same stage across the benchmark. We report no
paint metric, since there is no ground-truth material map and a colour
distance against a generated reference would score the reference rather than the model.

\runin{Articulation.}
Articulation, in contrast, can be checked objectively, since the physics simulator is an independent oracle. Across the
$18$ objects we have articulated and validated, $14$ clear every dynamics phase
-- settle, per-joint actuation, rest-contact scan, and URDF re-import -- and
$9$ of those also clear the advisory asset rules; the four that fail outright
are one asset that would not load and three whose actuators never traverse
(\cref{fig:articulation}).

\section{Limitations}
\label{sec:limitations}

\tether{} produces editable procedural assemblies that surpass native
generators on judged quality, yet three limitations remain. First, the agent perceives the build only through rendered images. A
rendered view conveys silhouette and placement well,
but occluded interiors and fine contact geometry can hide from every camera;
depth maps, cross-sections, or direct mesh queries would strengthen the
loop. Second, CSG captures mechanical, hard-surface
objects precisely, whereas organic, freeform surfaces are expressed less
naturally than in a neural field. Third, the mate vocabulary covers rigid joints only, and a mate the plan never declares cannot be verified.

\section{Conclusion}
\label{sec:conclusion}

We presented \tether{}, a procedural assembly agent that generates a 3D
object as the program that builds it, a parametric assembly whose named parts
are joined by typed, machine-checkable mates. A frozen LLM authors this representation with no 3D training, planning the
assembly graph, building the program part by part with placement solved from
the mated frames and every commit verified, and refining the result through
a decoupled vision critic. The same graph then carries per-part materials and a
simulator-validated articulation. On P3D-Bench and on MechBench-36, judged blind, \tether{}
leads every native generator and prior 3D-code agent on judged quality,
carries the sharpest edges of any method we measure, and is the only system
whose output re-opens as an editable, part-structured program. Its principles carry to program-based generation at large: casting shape
generation as coding, joining parts by verifiable mates, separating the
critic from the fixer, and validating every claim in a compiler or a
simulator.

{\small
\bibliographystyle{plain}
\clearpage
\bibliography{ref}

@inproceedings{poole2023dreamfusion,
  title     = {{DreamFusion}: Text-to-3D using 2D Diffusion},
  author    = {Poole, Ben and Jain, Ajay and Barron, Jonathan T. and Mildenhall, Ben},
  booktitle = {International Conference on Learning Representations (ICLR)},
  year      = {2023}
}

@inproceedings{lin2023magic3d,
  title     = {{Magic3D}: High-Resolution Text-to-3D Content Creation},
  author    = {Lin, Chen-Hsuan and Gao, Jun and Tang, Luming and Takikawa, Towaki and Zeng, Xiaohui and Huang, Xun and Kreis, Karsten and Fidler, Sanja and Liu, Ming-Yu and Lin, Tsung-Yi},
  booktitle = {IEEE/CVF Conference on Computer Vision and Pattern Recognition (CVPR)},
  year      = {2023}
}

@inproceedings{wang2023prolificdreamer,
  title     = {{ProlificDreamer}: High-Fidelity and Diverse Text-to-3D Generation with Variational Score Distillation},
  author    = {Wang, Zhengyi and Lu, Cheng and Wang, Yikai and Bao, Fan and Li, Chongxuan and Su, Hang and Zhu, Jun},
  booktitle = {Advances in Neural Information Processing Systems (NeurIPS)},
  year      = {2023}
}

@article{nichol2022point,
  title   = {{Point-E}: A System for Generating 3D Point Clouds from Complex Prompts},
  author  = {Nichol, Alex and Jun, Heewoo and Dhariwal, Prafulla and Mishkin, Pamela and Chen, Mark},
  journal = {arXiv preprint arXiv:2212.08751},
  year    = {2022}
}

@article{jun2023shape,
  title   = {{Shap-E}: Generating Conditional 3D Implicit Functions},
  author  = {Jun, Heewoo and Nichol, Alex},
  journal = {arXiv preprint arXiv:2305.02463},
  year    = {2023}
}

@inproceedings{zhao2023michelangelo,
  title     = {{Michelangelo}: Conditional 3D Shape Generation based on Shape-Image-Text Aligned Latent Representation},
  author    = {Zhao, Zibo and Liu, Wen and Chen, Xin and Zeng, Xianfang and Wang, Rui and Cheng, Pei and Fu, Bin and Chen, Tao and Yu, Gang and Gao, Shenghua},
  booktitle = {Advances in Neural Information Processing Systems (NeurIPS)},
  year      = {2023}
}

@inproceedings{wu2024direct3d,
  title     = {{Direct3D}: Scalable Image-to-3D Generation via 3D Latent Diffusion Transformer},
  author    = {Wu, Shuang and Lin, Youtian and Zhang, Feihu and Zeng, Yifei and Xu, Jingxi and Torr, Philip and Cao, Xun and Yao, Yao},
  booktitle = {Advances in Neural Information Processing Systems (NeurIPS)},
  year      = {2024}
}

@inproceedings{hong2024lrm,
  title     = {{LRM}: Large Reconstruction Model for Single Image to 3D},
  author    = {Hong, Yicong and Zhang, Kai and Gu, Jiuxiang and Bi, Sai and Zhou, Yang and Liu, Difan and Liu, Feng and Sunkavalli, Kalyan and Bui, Trung and Tan, Hao},
  booktitle = {International Conference on Learning Representations (ICLR)},
  year      = {2024}
}

@inproceedings{tang2024lgm,
  title     = {{LGM}: Large Multi-View Gaussian Model for High-Resolution 3D Content Creation},
  author    = {Tang, Jiaxiang and Chen, Zhaoxi and Chen, Xiaokang and Wang, Tengfei and Zeng, Gang and Liu, Ziwei},
  booktitle = {European Conference on Computer Vision (ECCV)},
  year      = {2024}
}

@inproceedings{sharma2018csgnet,
  title     = {{CSGNet}: Neural Shape Parser for Constructive Solid Geometry},
  author    = {Sharma, Gopal and Goyal, Rishabh and Liu, Difan and Kalogerakis, Evangelos and Maji, Subhransu},
  booktitle = {IEEE/CVF Conference on Computer Vision and Pattern Recognition (CVPR)},
  year      = {2018}
}

@inproceedings{kania2020ucsgnet,
  title     = {{UCSG-Net}: Unsupervised Discovering of Constructive Solid Geometry Tree},
  author    = {Kania, Kacper and Zi{\k{e}}ba, Maciej and Kajdanowicz, Tomasz},
  booktitle = {Advances in Neural Information Processing Systems (NeurIPS)},
  year      = {2020}
}

@article{jones2020shapeassembly,
  title   = {{ShapeAssembly}: Learning to Generate Programs for 3D Shape Structure Synthesis},
  author  = {Jones, R. Kenny and Barton, Theresa and Xu, Xianghao and Wang, Kai and Jiang, Ellen and Guerrero, Paul and Mitra, Niloy J. and Ritchie, Daniel},
  journal = {ACM Transactions on Graphics (SIGGRAPH Asia)},
  volume  = {39},
  number  = {6},
  year    = {2020}
}

@inproceedings{wu2021deepcad,
  title     = {{DeepCAD}: A Deep Generative Network for Computer-Aided Design Models},
  author    = {Wu, Rundi and Xiao, Chang and Zheng, Changxi},
  booktitle = {IEEE/CVF International Conference on Computer Vision (ICCV)},
  year      = {2021}
}

@article{willis2021fusion360,
  title   = {{Fusion 360 Gallery}: A Dataset and Environment for Programmatic CAD Construction from Human Design Sequences},
  author  = {Willis, Karl D.D. and Pu, Yewen and Luo, Jieliang and Chu, Hang and Du, Tao and Lambourne, Joseph G. and Solar-Lezama, Armando and Matusik, Wojciech},
  journal = {ACM Transactions on Graphics (SIGGRAPH)},
  volume  = {40},
  number  = {4},
  year    = {2021}
}

@inproceedings{xu2022skexgen,
  title     = {{SkexGen}: Autoregressive Generation of CAD Construction Sequences with Disentangled Codebooks},
  author    = {Xu, Xiang and Willis, Karl D.D. and Lambourne, Joseph G. and Cheng, Chin-Yi and Jayaraman, Pradeep Kumar and Furukawa, Yasutaka},
  booktitle = {International Conference on Machine Learning (ICML)},
  year      = {2022}
}

@inproceedings{hu2024scenecraft,
  title     = {{SceneCraft}: An {LLM} Agent for Synthesizing 3D Scenes as Blender Code},
  author    = {Hu, Ziniu and Iscen, Ahmet and Jain, Aashi and Kipf, Thomas and Yue, Yisong and Ross, David A. and Schmid, Cordelia and Fathi, Alireza},
  booktitle = {International Conference on Machine Learning (ICML)},
  year      = {2024}
}

@article{makatura2024scad,
  title   = {How Can Large Language Models Help Humans in Design and Manufacturing?},
  author  = {Makatura, Liane and Foshey, Michael and Wang, Bohan and H{\"a}hnlein, Felix and Ma, Pingchuan and Deng, Bolei and Tjandrasuwita, Megan and Spielberg, Andrew and Owens, Crystal Elaine and Chen, Peter Yichen and others},
  journal = {arXiv preprint arXiv:2307.14377},
  year    = {2023}
}

@inproceedings{khan2024text2cad,
  title     = {{Text2CAD}: Generating Sequential CAD Designs from Beginner-to-Expert Level Text Prompts},
  author    = {Khan, Mohammad Sadil and Sinha, Sankalp and Sheikh, Talha Uddin and Stricker, Didier and Ali, Sk Aziz and Afzal, Muhammad Zeshan},
  booktitle = {Advances in Neural Information Processing Systems (NeurIPS)},
  year      = {2024}
}

@inproceedings{yuan2024cadtalk,
  title     = {{CADTalk}: An Algorithm and Benchmark for Semantic Commenting of CAD Programs},
  author    = {Yuan, Haocheng and Xu, Jing and Pan, Hao and Bousseau, Adrien and Mitra, Niloy J. and Li, Changjian},
  booktitle = {IEEE/CVF Conference on Computer Vision and Pattern Recognition (CVPR)},
  year      = {2024}
}

@inproceedings{huang2024selfcorrect,
  title     = {Large Language Models Cannot Self-Correct Reasoning Yet},
  author    = {Huang, Jie and Chen, Xinyun and Mishra, Swaroop and Zheng, Huaixiu Steven and Yu, Adams Wei and Song, Xinying and Zhou, Denny},
  booktitle = {International Conference on Learning Representations (ICLR)},
  year      = {2024}
}

@inproceedings{madaan2023selfrefine,
  title     = {{Self-Refine}: Iterative Refinement with Self-Feedback},
  author    = {Madaan, Aman and Tandon, Niket and Gupta, Prakhar and Hallinan, Skyler and Gao, Luyu and Wiegreffe, Sarah and Alon, Uri and Dziri, Nouha and Prabhumoye, Shrimai and Yang, Yiming and others},
  booktitle = {Advances in Neural Information Processing Systems (NeurIPS)},
  year      = {2023}
}

@inproceedings{shinn2023reflexion,
  title     = {{Reflexion}: Language Agents with Verbal Reinforcement Learning},
  author    = {Shinn, Noah and Cassano, Federico and Berman, Edward and Gopinath, Ashwin and Narasimhan, Karthik and Yao, Shunyu},
  booktitle = {Advances in Neural Information Processing Systems (NeurIPS)},
  year      = {2023}
}

@inproceedings{wu2024gpteval3d,
  title     = {{GPT-4V(ision)} is a Human-Aligned Evaluator for Text-to-3D Generation},
  author    = {Wu, Tong and Yang, Guandao and Li, Zhibing and Zhang, Kai and Liu, Ziwei and Guibas, Leonidas and Lin, Dahua and Wetzstein, Gordon},
  booktitle = {IEEE/CVF Conference on Computer Vision and Pattern Recognition (CVPR)},
  year      = {2024}
}

@inproceedings{yang2023idea2img,
  title     = {{Idea2Img}: Iterative Self-Refinement with {GPT-4V} for Automatic Image Design and Generation},
  author    = {Yang, Zhengyuan and Wang, Jianfeng and Li, Linjie and Lin, Kevin and Lin, Chung-Ching and Liu, Zicheng and Wang, Lijuan},
  booktitle = {European Conference on Computer Vision (ECCV)},
  year      = {2024}
}

@article{attene2013repair,
  title   = {Polygon Mesh Repairing: An Application Perspective},
  author  = {Attene, Marco and Campen, Marcel and Kobbelt, Leif},
  journal = {ACM Computing Surveys},
  volume  = {45},
  number  = {2},
  year    = {2013}
}

@article{huang2020manifoldplus,
  title   = {{ManifoldPlus}: A Robust and Scalable Watertight Manifold Surface Generation Method for Triangle Soups},
  author  = {Huang, Jingwei and Zhou, Yichao and Guibas, Leonidas},
  journal = {arXiv preprint arXiv:2005.11621},
  year    = {2020}
}

@inproceedings{xiang2025trellis,
  title     = {{TRELLIS}: Structured 3D Latents for Scalable and Versatile 3D Generation},
  author    = {Xiang, Jianfeng and Lv, Zelong and Xu, Sicheng and Deng, Yu and Wang, Ruicheng and Zhang, Bowen and Chen, Dong and Tong, Xin and Yang, Jiaolong},
  booktitle = {IEEE/CVF Conference on Computer Vision and Pattern Recognition (CVPR)},
  year      = {2025}
}

@article{xiang2025trellis2,
  title   = {{TRELLIS.2}: Native and Compact Structured Latents for 3D Generation},
  author  = {Xiang, Jianfeng and Chen, Xiaoxue and Xu, Sicheng and Wang, Ruicheng and Lv, Zelong and Deng, Yu and Zhu, Hongyuan and Dong, Yue and Zhao, Hao and Yuan, Nicholas Jing and Yang, Jiaolong},
  journal = {arXiv preprint arXiv:2512.14692},
  year    = {2025}
}

@article{hunyuan3d2025,
  title   = {{Hunyuan3D 2.1}: From Images to High-Fidelity 3D Assets with Production-Ready {PBR} Material},
  author  = {{Team Hunyuan3D}},
  journal = {arXiv preprint arXiv:2506.15442},
  year    = {2025}
}

@article{jia2025ultrashape,
  title   = {{UltraShape} 1.0: High-Fidelity 3D Shape Generation via Scalable Geometric Refinement},
  author  = {Jia, Tanghui and Yan, Dongyu and Hao, Dehao and Li, Yang and Zhang, Kaiyi and He, Xianyi and Li, Lanjiong and Wang, Yuhan and Chen, Jinnan and Jiang, Lutao and Yin, Qishen and Quan, Long and Chen, Ying-Cong and Yuan, Li},
  journal = {arXiv preprint arXiv:2512.21185},
  year    = {2025}
}

@inproceedings{wu2025direct3ds2,
  title   = {{Direct3D-S2}: Gigascale 3D Generation Made Easy with Spatial Sparse Attention},
  author  = {Wu, Shuang and Lin, Youtian and Zhang, Feihu and Zeng, Yifei and Yang, Yikang and Bao, Yajie and Qian, Jiachen and Zhu, Siyu and Cao, Xun and Torr, Philip and Yao, Yao},
  booktitle = {Advances in Neural Information Processing Systems (NeurIPS)},
  year      = {2025}
}

@inproceedings{guan2025cadcoder,
  title     = {{CAD-Coder}: Text-to-CAD Generation with Chain-of-Thought and Geometric Reward},
  author    = {Guan, Yandong and Wang, Xilin and Xing, Ximing and Zhang, Jing and Xu, Dong and Yu, Qian},
  booktitle = {Advances in Neural Information Processing Systems (NeurIPS)},
  year      = {2025}
}

@inproceedings{kolodiazhnyi2026cadrille,
  title     = {cadrille: Multi-modal CAD Reconstruction with Reinforcement Learning},
  author    = {Kolodiazhnyi, Maksim and Tarasov, Denis and Zhemchuzhnikov, Dmitrii and Nikulin, Alexander and Zisman, Ilya and Vorontsova, Anna and Konushin, Anton and Kurenkov, Vladislav and Rukhovich, Danila},
  booktitle = {International Conference on Learning Representations (ICLR)},
  year      = {2026}
}

@article{shui2026articad,
  title   = {{ArtiCAD}: Articulated CAD Assembly Design via Multi-Agent Code Generation},
  author  = {Shui, Yuan and Guan, Yandong and Zhang, Zhanwei and Hu, Juncheng and Zhang, Jing and Xu, Dong and Yu, Qian},
  journal = {arXiv preprint arXiv:2604.10992},
  year    = {2026}
}

@article{zhou2026articraft,
  title   = {{ArtiCraft}: An Agentic System for Scalable Articulated 3D Asset Generation},
  author  = {Zhou, Matt and Li, Ruining and Lyu, Xiaoyang and Song, Zhaomou and Huang, Zhening and Zheng, Chuanxia and Rupprecht, Christian and Vedaldi, Andrea and Wu, Shangzhe},
  journal = {arXiv preprint arXiv:2605.15187},
  year    = {2026}
}

@article{caddesigner2025,
  title   = {{CADDesigner}: Conceptual CAD Model Generation with a General-Purpose Agent},
  author  = {Fan, Fengxiao and Ni, Jingzhe and Yin, Xiaolong and Wang, Sirui and Lu, Xingyu and Zou, Qiang and Tong, Ruofeng and Tang, Min and Du, Peng},
  journal = {arXiv preprint arXiv:2508.01031},
  year    = {2025}
}

@article{cadsmith2026,
  title   = {{CadSmith}: Multi-Agent CAD Generation with Programmatic Geometric Validation},
  author  = {Barkley, Jesse and Loghmani, Rumi and Barati Farimani, Amir},
  journal = {arXiv preprint arXiv:2603.26512},
  year    = {2026}
}

@misc{adamcad,
  author       = {{Adam CAD}},
  title        = {Adam: An {AI} CAD Copilot for Hardware Teams},
  year         = {2026},
  howpublished = {\url{https://adam.new/}},
  note         = {Commercial product; accessed 2026-07-19}
}

@article{p3dbench,
  title   = {{P3D-Bench}: Benchmarking {MLLMs} for Parametric 3D Generation and Structural Reasoning},
  author  = {Yang, Yikang and Hu, Zhanpeng and Lin, Youtian and Zhou, Mengqi and Xu, Jingxi and Zhang, Feihu and Liu, Jiaheng and Yao, Yao},
  journal = {arXiv preprint arXiv:2606.11152},
  year    = {2026}
}

@article{jones2021automate,
  title={AutoMate: A Dataset and Learned Model for Fixturing and Mating in {CAD} Assemblies},
  author={Jones, Benjamin and Hildreth, Dalton and Chen, Duowen and Baran, Ilya and Kim, Vladimir G. and Schulz, Adriana},
  journal={ACM Transactions on Graphics (SIGGRAPH Asia)},
  volume={40},
  number={6},
  year={2021}
}

@inproceedings{willis2022joinable,
  title={{JoinABLe}: Learning Bottom-up Assembly of Parametric {CAD} Joints},
  author={Willis, Karl D. D. and Jayaraman, Pradeep Kumar and Chu, Hang and Tian, Yunsheng and Li, Yifei and Grandi, Daniele and Sanghi, Aditya and Tran, Linh and Lambourne, Joseph G. and Solar-Lezama, Armando and Matusik, Wojciech},
  booktitle={IEEE/CVF Conference on Computer Vision and Pattern Recognition (CVPR)},
  year={2022}
}

@inproceedings{deitke2022procthor,
  title={{ProcTHOR}: Large-Scale Embodied {AI} Using Procedural Generation},
  author={Deitke, Matt and VanderBilt, Eli and Herrasti, Alvaro and Weihs, Luca and Salvador, Jordi and Ehsani, Kiana and Han, Winson and Kolve, Eric and Farhadi, Ali and Kembhavi, Aniruddha and Mottaghi, Roozbeh},
  booktitle={Advances in Neural Information Processing Systems (NeurIPS)},
  year={2022}
}

@inproceedings{yang2024holodeck,
  title={Holodeck: Language Guided Generation of {3D} Embodied {AI} Environments},
  author={Yang, Yue and Sun, Fan-Yun and Weihs, Luca and VanderBilt, Eli and Herrasti, Alvaro and Han, Winson and Wu, Jiajun and Haber, Nick and Krishna, Ranjay and Liu, Lingjie and Callison-Burch, Chris and Yatskar, Mark and Kembhavi, Aniruddha and Clark, Christopher},
  booktitle={IEEE/CVF Conference on Computer Vision and Pattern Recognition (CVPR)},
  year={2024}
}

@inproceedings{feng2023layoutgpt,
  title={{LayoutGPT}: Compositional Visual Planning and Generation with Large Language Models},
  author={Feng, Weixi and Zhu, Wanrong and Fu, Tsu-Jui and Jampani, Varun and Akula, Arjun and He, Xuehai and Basu, Sugato and Wang, Xin Eric and Wang, William Yang},
  booktitle={Advances in Neural Information Processing Systems (NeurIPS)},
  year={2023}
}

@inproceedings{raistrick2023infinigen,
  title={Infinite Photorealistic Worlds Using Procedural Generation},
  author={Raistrick, Alexander and Lipson, Lahav and Ma, Zeyu and Mei, Lingjie and Wang, Mingzhe and Zuo, Yiming and Kayan, Karhan and Wen, Hongyu and Han, Beining and Wang, Yihan and Newell, Alejandro and Law, Hei and Goyal, Ankit and Yang, Kaiyu and Deng, Jia},
  booktitle={IEEE/CVF Conference on Computer Vision and Pattern Recognition (CVPR)},
  year={2023}
}

@inproceedings{le2024articulate,
  title={Articulate-Anything: Automatic Modeling of Articulated Objects via a Vision-Language Foundation Model},
  author={Le, Long and Xie, Jason and Liang, William and Wang, Hung-Ju and Yang, Yue and Ma, Yecheng Jason and Vedder, Kyle and Krishna, Arjun and Jayaraman, Dinesh and Eaton, Eric},
  booktitle={International Conference on Learning Representations (ICLR)},
  year={2025}
}

@book{boothroyd2010dfma,
  title={Product Design for Manufacture and Assembly},
  author={Boothroyd, Geoffrey and Dewhurst, Peter and Knight, Winston A.},
  edition={3rd},
  publisher={CRC Press},
  year={2010}
}

@inproceedings{nash2020polygen,
  author    = {Nash, Charlie and Ganin, Yaroslav and Eslami, S. M. Ali and Battaglia, Peter W.},
  title     = {{PolyGen}: An Autoregressive Generative Model of 3D Meshes},
  booktitle = {International Conference on Machine Learning (ICML)},
  year      = {2020}
}

@inproceedings{siddiqui2024meshgpt,
  author    = {Siddiqui, Yawar and Alliegro, Antonio and Artemov, Alexey and Tommasi, Tatiana and Sirigatti, Daniele and Rosov, Vladislav and Dai, Angela and Nie{\ss}ner, Matthias},
  title     = {{MeshGPT}: Generating Triangle Meshes with Decoder-Only Transformers},
  booktitle = {IEEE/CVF Conference on Computer Vision and Pattern Recognition (CVPR)},
  year      = {2024}
}

@inproceedings{chen2024meshanything,
  author    = {Chen, Yiwen and He, Tong and Huang, Di and Ye, Weicai and Chen, Sijin and Tang, Jiaxiang and Chen, Xin and Cai, Zhongang and Yang, Lei and Yu, Gang and Lin, Guosheng and Zhang, Chi},
  title     = {{MeshAnything}: Artist-Created Mesh Generation with Autoregressive Transformers},
  booktitle = {International Conference on Learning Representations (ICLR)},
  year      = {2025}
}

@article{yang2024sampart3d,
  author  = {Yang, Yunhan and Huang, Yukun and Guo, Yuan-Chen and Lu, Liangjun and Wu, Xiaoyang and Lam, Edmund Y. and Cao, Yan-Pei and Liu, Xihui},
  title   = {{SAMPart3D}: Segment Any Part in 3D Objects},
  journal = {arXiv preprint arXiv:2411.07184},
  year    = {2024}
}

@article{liu2025partfield,
  author  = {Liu, Minghua and Uy, Mikaela Angelina and Xiang, Donglai and Su, Hao and Fidler, Sanja and Sharp, Nicholas and Gao, Jun},
  title   = {{PartField}: Learning 3D Feature Fields for Part Segmentation and Beyond},
  journal = {arXiv preprint arXiv:2504.11451},
  year    = {2025}
}

@article{yang2025holopart,
  author  = {Yang, Yunhan and Guo, Yuan-Chen and Huang, Yukun and Zou, Zi-Xin and Yu, Zhipeng and Li, Yangguang and Cao, Yan-Pei and Liu, Xihui},
  title   = {{HoloPart}: Generative 3D Part Amodal Segmentation},
  journal = {arXiv preprint arXiv:2504.07943},
  year    = {2025}
}

@article{openai2023gpt4,
  author  = {{OpenAI}},
  title   = {{GPT-4} Technical Report},
  journal = {arXiv preprint arXiv:2303.08774},
  year    = {2023}
}

@article{gemini2023,
  author  = {{Gemini Team}},
  title   = {Gemini: A Family of Highly Capable Multimodal Models},
  journal = {arXiv preprint arXiv:2312.11805},
  year    = {2023}
}
}

\end{document}